%% file: main.tex
\documentclass[sigconf,9pt,nonacm]{acmart}
\setcopyright{none}
\usepackage{placeins}
\usepackage{amsmath,booktabs,graphicx,tabularx}
\usepackage{multirow}
\usepackage{float}
\usepackage{xcolor}
\definecolor{pendingcolor}{RGB}{150,72,25}
\newcommand{\framework}{\textbf{BeFit}}
\newcommand{\rd}[2]{\begin{tabular}[t]{@{}r@{}}#1\\[-2pt]{($\downarrow$#2$\times$)}\end{tabular}}

\newcommand{\pp}{\,pp}

\title[BeFit]{BeFit: Shifting Complexity from Weights to Bernstein Activations for Resource-Constrained Sensing}

\author{Malak Gamal El-Din, Yasser Shoukry, Salma Elmalaki}
\affiliation{\institution{}\city{}\country{}}

\usepackage{pgfplots}\pgfplotsset{compat=1.18}
\usepackage[all]{nowidow}

\begin{document}
\input{sections/01_abstract}

\maketitle
\begingroup\renewcommand\thefootnote{}\footnotetext{All authors are with the
Department of Electrical Engineering and Computer Science, University of
California, Irvine, CA 92697 USA (e-mail: mgamalel@uci.edu;
yshoukry@uci.edu; selmalak@uci.edu).}\endgroup

\input{sections/03_introduction}
\input{sections/04_background_related}
\input{sections/05_framework}
\input{sections/06_nas}
\input{sections/07_distillation}
\input{sections/08_projection}
\input{sections/10_kernels}
\input{sections/11_extensions}
\input{sections/12_results_setup}
\input{sections/13_results_pipeline}
\input{sections/14_results_nas}
\input{sections/15_results_kd}
\input{sections/16_results_projection}
\input{sections/17_results_kernels}
\input{sections/18_results_transformer_mlp}
\input{sections/20_discussion}
\input{sections/21_conclusion}

\begin{acks}
This material is based upon work supported by the U.S National Science Foundation (NSF) under Grant No.~2504809, and by the NSF Graduate Research Fellowship Program under Grant No. DGE-2235784. Any opinions, findings, and conclusions or recommendations expressed in this material are those of the author(s) and do not necessarily reflect the views of the NSF.
\end{acks}

\raggedbottom
\bibliographystyle{ACM-Reference-Format}
\bibliography{references}

\appendix
\input{sections/22_appendix}
\end{document}

%% file: sections/01_abstract.tex
\begin{abstract}
Time-series sensing increasingly relies on neural networks, yet deployment on resource-constrained microcontrollers remains challenging. Aggressive model compression can reduce accuracy, while restricting weights to multiplication-free representations introduces further losses.
We present \framework{}, an end-to-end framework that shifts model capacity from
weights to learnable Bernstein polynomial activations. \framework{} (i)~finds
compact Bernstein-activated convolutional neural networks (BeCNNs) through hardware-aware neural architecture search and trains them via knowledge
distillation; (ii)~projects their weights to multiplication-free codes,
freezes the codes, and recovers the resulting accuracy loss through the learnable activations; and (iii)~generates
model-specific kernels that exploit the projected weights for multiplication-free convolutions. On human activity recognition, BeCNNs are 2.75 percentage points (pp) more accurate on average than ReLU CNNs of identical architecture. On a TI CC3220SF (Cortex-M4), weight projection speeds BeCNNs up by 2.2$\times$ on average while losing 0.8\,pp, compared with 2.4\,pp for ReLU CNNs. Against the MLPerf Tiny DS-CNN reference model running on TensorFlow Lite Micro with CMSIS-NN, the unprojected BeCNN achieves 1.0\,pp higher
accuracy, 2.2$\times$ lower latency, and 6.4$\times$ lower flash, while the projected model remains within 1.0\,pp of the reference accuracy and runs 5.9$\times$ faster. In emulation on a RISC-V core without a hardware multiplier, the instruction-count reduction relative to the DS-CNN increases from 5.2$\times$ on the Cortex-M4 to 147$\times$. 
\framework{} also extends to transformers and MLPs, and across all tasks the final models stay within 1.56\,pp of their float32 teachers on average.
\end{abstract}

%% file: sections/03_introduction.tex
\section{Introduction}
\label{sec:intro}
Sensing applications such as human activity recognition (HAR) on wearables and keyword spotting (KWS) run neural networks continuously and close to the sensor, often on low-power microcontrollers (MCUs) with a few hundred kilobytes of memory and limited compute. Fitting these budgets requires shrinking the model and reducing costly multiplications, both of which lower accuracy. Recovering that accuracy within MCU budgets remains a challenge.

Existing work addresses parts of this problem. Runtimes and kernel libraries~\cite{tflm,cmsisnn}, hardware-aware architecture searches~\cite{mcunet,micronets} and memory co-design~\cite{vmcu} fit models to the MCU, but keep fixed activations and weights that must be multiplied. Ternary and power-of-two quantization~\cite{twn,inq} removes the multiplications but keeps fixed activations, so the restricted weights must recover the lost accuracy on their own, which is hard in compact models. Existing MCU work on learnable activations, such as TinyKAN-HAR~\cite{lamaakal2026tinykan}, keeps weights that still multiply. None of these approaches combines frozen multiplication-free weights with learnable polynomial activations to recover the resulting accuracy loss.

We present \framework{}, an end-to-end framework that turns a large pretrained network into a compact MCU model with multiplication-free convolutions. \framework{} replaces a CNN's conventional activations with learnable Bernstein polynomials~\cite{bernstein1912,deepbern}, forming a BeCNN. Next, it finds a compact BeCNN with a hardware-aware architecture search and trains it via knowledge distillation (KD). It then projects the convolutional weights onto a multiplication-free set, freezes them and recovers the lost accuracy through the activations, and finally generates kernels that exploit the projected weights for deployment.

To our knowledge, \framework{} is the first to use learnable Bernstein activations for sensing on MCUs, and the first to use them to recover the accuracy lost when weights are projected onto a multiplication-free set.

The contributions of this work are as follows:
\begin{itemize}
\item \textbf{An end-to-end framework for MCU deployment.} \framework{}, built around learnable Bernstein activations, turns a large pretrained network into an integer MCU image through hardware-aware architecture search, knowledge distillation, weight projection with recovery, and kernel generation with lookup-table activations.
\item \textbf{BeCNNs: Bernstein activations for sensing on MCUs.} We bring learnable Bernstein activations to compact CNNs for sensing on MCUs, replacing their conventional activations to form Bernstein-activated CNNs (BeCNNs). They restore the expressiveness compact models lose under strong compression, at almost no additional inference cost.
\item \textbf{Weight projection with activation recovery.} We project the convolutional weights to multiplication-free codes and freeze them, then exploit the learnable Bernstein activations to recover the model's lost accuracy.
\item \textbf{Projection-aware kernel generation.} \framework{} generates architecture-specific kernels that exploit the projected weights for multiplication-free convolutions, and compiles each activation into a lookup table for efficient inference.
\item \textbf{Evaluation on six tasks and a real MCU.} We show that BeCNNs are more accurate than ReLU CNNs under strong compression and lose little accuracy after projection. We evaluate \framework{} end to end on the TI CC3220SF~\cite{cc3220sf} against TensorFlow Lite Micro~\cite{tflm} with CMSIS-NN~\cite{cmsisnn} and TinyEngine~\cite{mcunet}. We also test it in emulation on RISC-V with no hardware multiplier, and extend it to transformers and multilayer perceptrons (MLPs).
\end{itemize}

%% file: sections/04_background_related.tex
\section{Background and Related Work}
\label{sec:background}\label{sec:related}

\paragraph{\textbf{MCU frameworks.}}
\label{sec:related-frameworks}
TensorFlow Lite Micro (TFLM)~\cite{tflm} is an open-source MCU runtime used by MLPerf Tiny’s reference implementations~\cite{mlperftiny}. Its interpreter executes a quantized model graph and calls a kernel per operator, taken from CMSIS-NN~\cite{cmsisnn} on Arm Cortex-M. While TFLM runs a given model, hardware-aware architecture searches design models that fit the MCU: MCUNet~\cite{mcunet} co-designs the search with its own inference engine, and MicroNAS~\cite{micronas} targets time-series classifiers under memory and latency limits. However, these works keep standard layers with fixed activations and multiplied weights. \framework{} instead changes the building blocks themselves, with learnable Bernstein activations and multiplication-free weights, and builds the search, training and kernels around them.

\paragraph{\textbf{Learnable activations.}}
\label{sec:related-activations}
Padé Activation Units~\cite{pau} learn rational activations end to end, over an unbounded input range. Kolmogorov--Arnold Networks (KANs)~\cite{kan} replace each weight of a multilayer perceptron with a learnable one-dimensional function on its edge, and these functions can be compiled into lookup tables on MCUs~\cite{kuznetsov2026lutkan}. TinyKAN-HAR~\cite{lamaakal2026tinykan} applies one learnable spline per neuron after each linear layer of a fully connected network, stored as a lookup table, and keeps int8 weights; it targets zero-shot recognition and explainability. Faggi et al.~\cite{faggi2026cnnkan} pair a CNN backbone with a KAN classifier head, but the convolutions keep fixed activations. \framework{} instead places one learnable Bernstein curve per filter or layer inside a standard CNN, compiles each into a lookup table, and uses these curves to recover the accuracy lost to compression and multiplication-free weights.

\paragraph{\textbf{Bernstein polynomial networks}}
\label{sec:bernstein}
Khedr and Shoukry~\cite{deepbern} introduce DeepBern-Nets, feed-forward networks in which standard activations are replaced by learnable Bernstein polynomials, and use their coefficient bounds to certify image classifiers. Albool et al.~\cite{albool2026bern} prove that their approximation error decays exponentially in the network depth, against a polynomial rate for ReLU networks. Bern2Edge~\cite{bern2edge} later distills feed-forward networks into Bernstein networks for FPGA deployment, either as lookup tables or as symbolic rules. \framework{} instead brings Bernstein-activated CNNs to time-series sensing on resource-constrained MCUs, and exploits them to recover the accuracy lost to multiplication-free convolutions.

\paragraph{\textbf{Multiplication-free weights.}}
\label{sec:related-projection}
Ternary ($\{0,\pm1\}$) and power-of-two ($\{0,\pm2^k\}$) weights need no multiplier. Ternary Weight Networks (TWN)~\cite{twn} approximate each filter by a scale times ternary codes and train latent full-precision weights through the quantizer, and INQ~\cite{inq} quantizes weights to powers of two or zero in steps, retraining the remaining weights after each step. DeepShift~\cite{deepshift} trains power-of-two weights so that each multiply becomes a shift and a sign flip, and ShiftCNN~\cite{shiftcnn} converts pretrained CNNs to power-of-two weights for custom accelerators. All keep fixed activations and recover accuracy, when needed, through the weights. ReActNet~\cite{liu2020reactnet} binarizes weights and activations and adds learnable shifts and slopes to the sign and PReLU functions, but still trains the weights. \framework{} instead freezes the projected codes and recovers accuracy through the learnable Bernstein curves. To our knowledge, it is the first to do so with learnable polynomial activations.

Sensing systems have also cut multiplications by leaving standard networks behind: Bakar et al.~\cite{bakar2022tsetlin} run Tsetlin machines on batteryless MCUs, whose inference evaluates Boolean clauses and sums their votes without any multiplication, and Fast-Inf~\cite{fastinf} reduces the multiply count with binary tree-based networks whose cost grows logarithmically with network size. Both, however, give up the CNN. \framework{} instead keeps the convolutional backbone with multiplication-free weights.

\paragraph{\textbf{Kernel generation.}}
\label{sec:related-kernels}
Generated kernels are usually specialized to the model's structure. TinyEngine in MCUNet~\cite{mcunet} generates kernels and memory schedules for each layer's shape, and Lupe~\cite{lupe} converts network descriptions into code for ultra-low-power MCUs, but both still load and multiply the weights. Other work specializes the code to the weight values. On FPGAs, unrolled ternary networks~\cite{tridgell2019unrolling} store the weights in the routing of pruned adder trees. On MCUs, ATAMAN~\cite{ataman} writes int8 weights into generated Cortex-M7 kernels, but still multiplies, and Neuro-C~\cite{neuroc} fixes ternary connections on a Cortex-M0, but has no convolutions. In contrast, \framework{} generates multiplication-free convolution kernels for MCUs directly from its projected weights (Section~\ref{sec:kernels}).

%% file: sections/05_framework.tex
\section{BeFit Framework}
\label{sec:framework}
\begin{figure*}[t]\centering
\includegraphics[width=\textwidth,trim={0 4cm 0 0},clip]{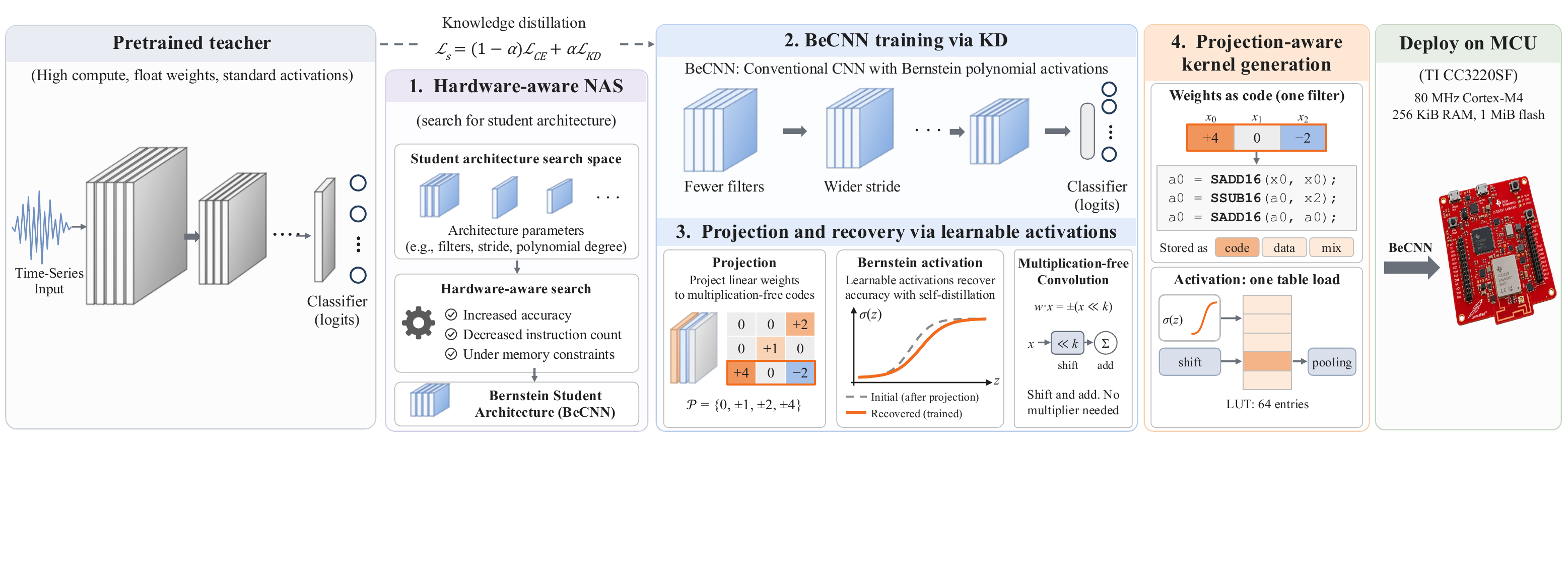}
\Description{Five stages from left to right: a pretrained teacher; a hardware-aware NAS that searches for a compact student for higher accuracy and lower instruction count under memory constraints; the compressed student BeCNN with fewer filters and wider strides, trained by distillation and quantized to int8, whose weights are projected to multiplication-free codes while learnable Bernstein activations recover accuracy with self-distillation, shown as one curve before and after recovery; projection-aware generated kernels, where one filter's codes become add and subtract instructions, stored as code, data or a mix, and each activation is one lookup-table load; and deployment on the TI CC3220SF, an 80 MHz Cortex-M4 with 256 KiB RAM and 1 MiB flash. Knowledge distillation links the teacher and the student.}
\caption{\framework{} pipeline. A hardware-aware NAS finds a compact BeCNN, and a high-accuracy teacher helps train it via KD. The BeCNN's convolutional weights are projected to multiplication-free codes, the learnable activations recover the lost accuracy, and a generated kernel exploits these codes for MCU deployment. Weights and curves are illustrative.}
\label{fig:framework}\end{figure*}

\framework{} turns a large pretrained teacher into a compact integer model for MCUs. Its central idea is to shift model capacity from the weights to the activations: the weights are made fewer and then multiplication-free, while learnable Bernstein activations carry the expressiveness the weights give up. This trade suits MCUs because the activations are cheap where the weights are not. Each activation curve compiles into a fixed-size lookup table, so its inference cost is small and independent of its degree. The building block is the BeCNN (Section~\ref{sec:becnn}), a standard CNN whose fixed activations are replaced by learnable Bernstein curves (Fig.~\ref{fig:becnn}). Around it, \framework{} runs four modules (Fig.~\ref{fig:framework}).

(1) A \textbf{hardware-aware NAS} (Section~\ref{sec:nas}) finds a compact BeCNN under the MCU's memory limits, trading accuracy against instructions measured on the target's own build. Because the polynomial degree does not change these costs, the search treats it as a free dimension and can raise the curves' expressiveness at no additional deployed inference cost, although training cost grows (Section~\ref{sec:setup}). (2) \textbf{Knowledge distillation} (Section~\ref{sec:compression}) then trains the selected BeCNN in full from the teacher. (3) \textbf{Projection and recovery} (Section~\ref{sec:projection}) remove the convolutions' multiplications: the convolutional weights are projected onto multiplication-free codes and frozen, and the accuracy lost to projection is recovered by retraining the Bernstein curves rather than the weights. (4) \textbf{Projection-aware kernel generation} (Section~\ref{sec:kernels}) specializes the inference kernels to the multiplication-free codes. Each Bernstein activation becomes a single table load, and requantization a shift. The result is an integer model whose convolutions need no multiplier. Section~\ref{sec:extensions} extends the same pipeline to transformers and MLPs.

\subsection{BeCNNs: Bernstein Convolutional Networks}
\label{sec:becnn}
\textbf{Bernstein polynomials.} A Bernstein polynomial $\sigma(z)$ of degree $N$, defined on the interval $[l, u]$, is written as~\cite{bernstein1912}:
\begin{equation}
\label{eq:bern}
  \sigma(z;\, l, u, \mathbf{c}) = \sum_{i=0}^{N} c_i \, B_{N,i}(t),
  \quad t = \frac{z - l}{u - l} \in [0,1], \quad z \in [l, u],
\end{equation}
where $z$ is the input to the polynomial function, $t$ is the normalized input, and $\mathbf{c} = (c_0, \ldots, c_N)$ are the polynomial coefficients. The Bernstein basis functions are defined as:
\begin{equation}
  B_{N,i}(t) = \binom{N}{i} t^i (1 - t)^{N-i}, \quad i = 0, \ldots, N.
\end{equation}
They are nonnegative and sum to one on $[0,1]$. Hence, the polynomial is a convex combination of its
coefficients over a bounded domain, with the coefficients directly controlling the activation shape.

\begin{figure*}[t]\centering\includegraphics[width=\textwidth,trim={0.65cm 3cm 0.65cm 0},clip]{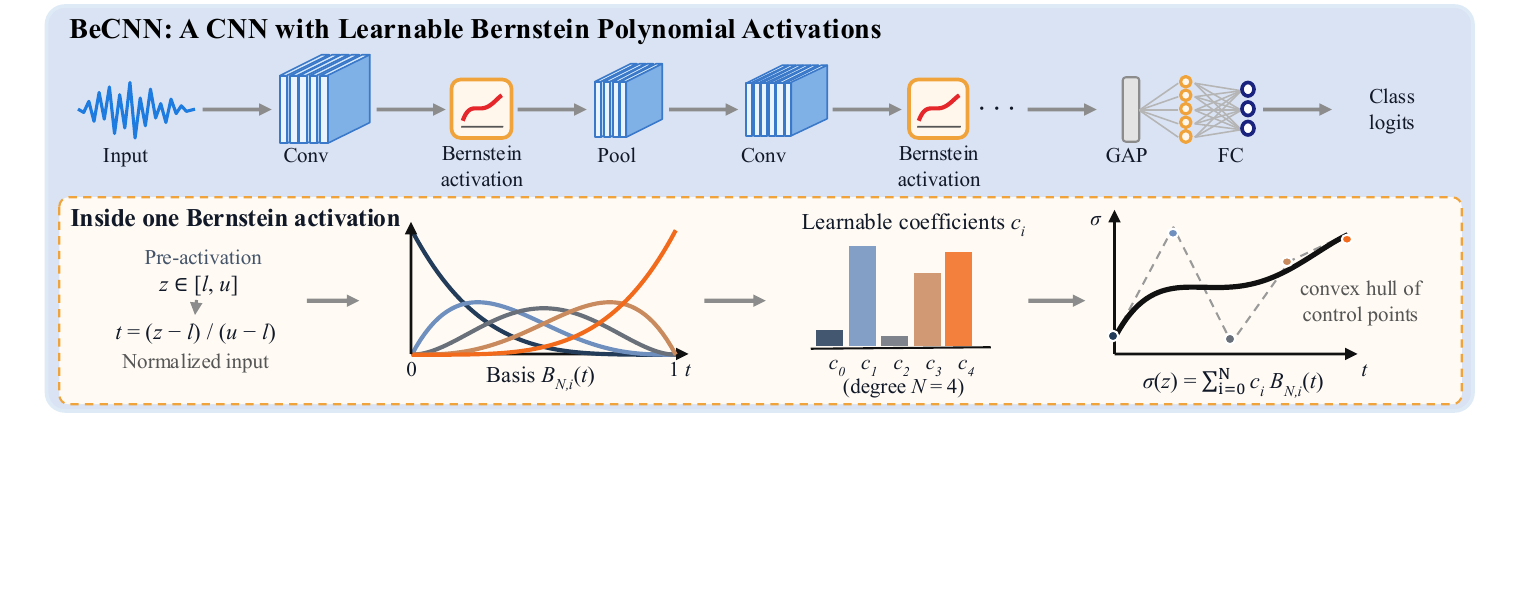}
\Description{Top: a CNN pipeline from an input signal through convolution, Bernstein activation, pooling, convolution, Bernstein activation, global average pooling and a fully connected layer to class logits. Bottom, inside one Bernstein activation: the pre-activation is normalized to t in [0,1], evaluated on five degree-4 Bernstein basis functions, weighted by five learnable coefficients, and summed into a smooth activation curve that lies within the convex hull of its control points.}

\caption{A BeCNN: a CNN whose activations are learnable Bernstein polynomials (Eq.~\eqref{eq:bern}). Each activation normalizes its input to $t\in[0,1]$ and outputs $\sum_i c_i B_{N,i}(t)$; only the coefficients are learned, and the curve stays within the convex hull of its control points. Layer count is schematic.}\label{fig:becnn}\end{figure*}

\textbf{BeCNNs.} \framework{} replaces a CNN's fixed activations with learnable Bernstein polynomials, forming a BeCNN (Fig.~\ref{fig:becnn}). A BeCNN keeps the convolutions, pooling and classifier of a standard CNN and learns one curve per filter, shared across all of the filter's positions.

\textbf{Per-filter Activations.} At each position, filter $j$ computes its pre-activation $z=\mathbf{w}_j^\top\mathbf{x}+b_j$ from its weights $\mathbf{w}_j$, bias $b_j$ and input $\mathbf{x}$, as in a standard CNN. The activation then normalizes $z$ to $[0,1]$ and evaluates the filter's curve (Eq.~\eqref{eq:bern}), with the filter's own bounds $l_j,u_j$ and learned coefficients $\mathbf{c}_j$.

\textbf{Why Bernstein activations.} Two properties suit Bernstein activations to resource-constrained MCUs. (1)~Parameter efficiency: for Lipschitz targets, the approximation error of a Bernstein network of degree $N$ and depth $D$ decays as $O(N^{-D})$, exponentially in the depth, against a polynomial rate for ReLU networks, and most tested feed-forward architectures need far fewer parameters than their ReLU equivalents~\cite{albool2026bern}. (2)~Deployability: each activation's normalized input lies in the fixed range $[0,1]$, so its curve can be precomputed offline into a lookup table whose size does not depend on the degree~\cite{bern2edge}.

%% file: sections/06_nas.tex
\subsection{Hardware-Aware Neural Architecture Search (NAS)}
\label{sec:nas}
The first step of \framework{} (Fig.~\ref{fig:framework}) is a hardware-aware search for the right BeCNN. In a conventional CNN with fixed activations, added capacity costs instructions. In contrast, a BeCNN can increase its representational capacity by raising the Bernstein polynomial degree without increasing deployed inference cost. The search therefore chooses the architecture and the polynomial degree together, trading accuracy against computation under the target MCU's memory limits.

\textbf{Search objective.} For an architecture $a$ with Bernstein degree $N$ and training hyperparameters $h$, let $\mathrm{Acc}_{\mathrm{val}}(a,N,h)$ be the validation accuracy of the trained int8 model and $I(a)$ its instructions per inference, before projection. The search approximates the Pareto set of
\begin{equation}
 \max_{a,N,h}\mathrm{Acc}_{\mathrm{val}}(a,N,h),\ \ \min_{a}I(a)\ \ \text{s.t.}\ F(a)\le F_{\max},\ R(a)\le R_{\max},
 \label{eq:nas}
\end{equation}
where $F$ is flash (model data plus inference code), $R$ is RAM (static data plus peak stack), and $F_{\max}$ and $R_{\max}$ are the MCU's limits. $I$, $F$ and $R$ come from a cost table measured before the search: every architecture is compiled with the target's build flags and run in the QEMU simulator with random weights, which is valid because the instruction count of the unprojected int8 model does not depend on its weight values.

\textbf{Polynomial degree.} The degree $N$ enters the accuracy objective of Eq.~\eqref{eq:nas} but none of its costs, since every curve compiles into the same lookup table whatever its degree (Section~\ref{sec:lut}). The search can therefore raise a curve's expressiveness at no additional deployed inference cost (only training cost grows), and it samples $N$ jointly with the architecture, so each shape gets the degree that suits it.

\textbf{Cheap training.} To estimate how an architecture performs, each trial trains from scratch on labels only, without distillation, with one seed and for a short, fixed number of epochs, and is scored on its int8 model with every curve compiled to its deployed lookup table, so the score includes the loss from tabulating the curves.

\textbf{Optimizer.} We use the multi-objective tree-structured Parzen estimator~\cite{bergstra2011tpe,ozaki2020motpe} in Optuna~\cite{akiba2019optuna}. Flash and RAM are trial constraints: an architecture that violates them is not trained, and its recorded violation steers the sampler away. Multi-objective asynchronous successive halving~\cite{li2020asha,schmucker2021moasha} prunes weak trials early. It ranks trials on both accuracy and instructions, so that cheap models stay on the Pareto front instead of being pruned for lower accuracy.

\textbf{Selecting students.} Because each trial uses one seed, the candidates of non-dominated rank $\le2$ (the Pareto front and the next front behind it) are retrained with two more seeds, and the front is taken on the three-seed validation mean. Under an instruction budget $I_{\max}$, we keep the candidate with the best mean among those with $I(a)\le I_{\max}$. This BeCNN becomes the student trained in \framework{}'s next step (Section~\ref{sec:compression}). Section~\ref{sec:exp-nas} gives the search spaces, the setup, the ReLU baseline and the results.

%% file: sections/07_distillation.tex
\subsection{BeCNN Training via Knowledge Distillation}
\label{sec:compression}
The second step of \framework{} (Fig.~\ref{fig:framework}) trains the student BeCNN found by the NAS. Since the search only trains each candidate briefly and on labels alone, knowledge distillation (KD) from the pretrained teacher now trains the student in full, before any weight projection.

\textbf{Knowledge distillation loss.} The student objective follows the standard distillation loss~\cite{hinton2015}:
\begin{equation}
 \mathcal L_s=(1-\alpha)\mathcal L_{\mathrm{CE}}+\alpha T^2\,\mathrm{KL}\big(S(y^*/T)\,\|\,S(\hat y/T)\big),
 \label{eq:kd}
\end{equation}
where $\mathcal L_{\mathrm{CE}}$ is the cross-entropy loss between student predictions and ground-truth labels, $S(\cdot)$ is the softmax, $\mathrm{KL}(\cdot\|\cdot)$ is the Kullback--Leibler divergence between the softened teacher and student distributions, and $\hat y$ and $y^*$ are the student and teacher logits. $T$ is the temperature controlling how soft the outputs are, and $\alpha$ balances supervision between hard labels and teacher guidance. The teacher stays frozen, sees the same batches as the student, and is discarded at inference.

\textbf{Bernstein bounds.} A Bernstein polynomial is evaluated on $[0,1]$, while a filter's pre-activations can take any value, so each activation needs bounds $[l_j,u_j]$ that map its pre-activation onto this interval (Eq.~\eqref{eq:bern}). BeCNN training starts with a short warm-up using fixed bounds. Each filter's bounds are then calibrated from the 0.1st and 99.9th percentiles of its pre-activations on the training data and stay fixed for the rest of training~\cite{bern2edge}. Fixing the bounds stabilizes training and gives each lookup table a known input range (Section~\ref{sec:lut}). Models with BatchNorm before the activation skip the calibration and keep the warm-up bounds.

\textbf{Out-of-bounds penalty.} Pre-activations outside the bounds $[l_j,u_j]$ are clamped, so the BeCNN adds an out-of-bounds penalty to its loss to discourage them:
\begin{equation}
 \mathcal L_{\mathrm{BeCNN}}=\mathcal L_s+\lambda_{\mathrm{ob}}\,\mathbb E\big[\max(0,- t)+\max(0,t-1)\big],
 \label{eq:ob}
\end{equation}
where $t$ is the normalized pre-activation of Eq.~\eqref{eq:bern} before clamping.

The result is a compact, high-accuracy BeCNN, which we refer to as \framework{}$_\text{dense}$. These dense models are the starting point of weight projection (Section~\ref{sec:projection}). Section~\ref{sec:exp-kd} compares them against equivalent ReLU networks under the same hardware budgets.

%% file: sections/08_projection.tex
\subsection{Projection and Recovery via Learnable Activations}
\label{sec:projection}
\begin{figure*}[t]\centering\includegraphics[width=\textwidth]{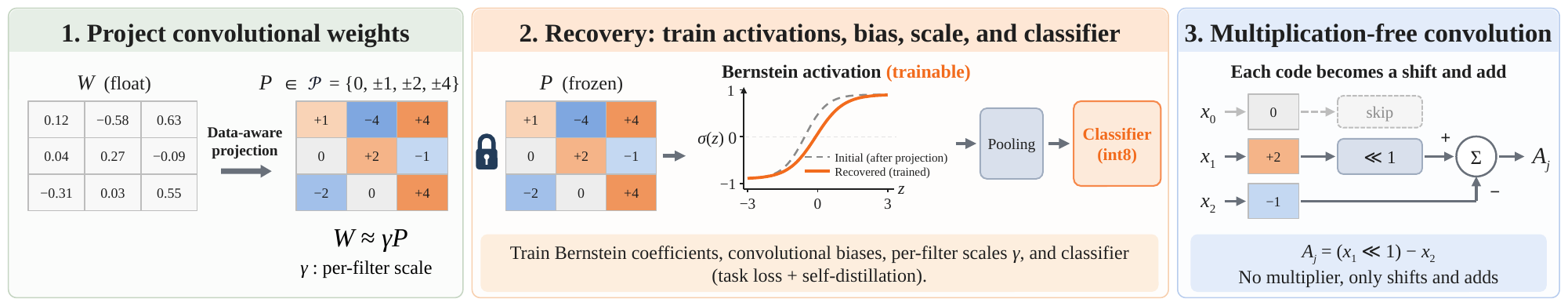}
\Description{Three panels. Left: data-aware projection of floating-point convolutional weights to power-of-two codes in {0, plus or minus 1, 2, 4}, times a per-filter scale. Middle: recovery with the codes frozen, training the Bernstein activation, convolutional biases, per-filter scales and the classifier with the task loss and self-distillation; the activation curve is shown after projection and after recovery, followed by pooling and an int8 classifier. Right: multiplication-free convolution, where each code becomes a shift and add: a zero code is skipped, a code of plus 2 is a left shift by one and a code of minus 1 is a subtraction.}
\caption{Projection, recovery and multiplication-free convolution. Convolutional weights are approximated as $W\approx\gamma P$ with a per-filter scale $\gamma$ and codes $P$. Recovery freezes $P$ and trains the Bernstein coefficients, biases, scales and classifier. At inference, nonzero codes are implemented using additions and shifts, so the convolution needs no multiplier. Matrices and curves are illustrative.}\label{fig:projection}\end{figure*}
The third step of \framework{} (Fig.~\ref{fig:framework}) projects the trained BeCNN's dense convolutional weights to multiplication-free codes and recovers the lost accuracy through the learnable Bernstein activations (Fig.~\ref{fig:projection}).

\subsubsection{\textbf{Projection to multiplication-free codes.}} Projection restricts each filter's weights to a small set of values that need no multiplication. We first write the weights of filter $j$ as a vector $\mathbf{w}_j$ and replace it with $\gamma_j\mathbf{p}_j$: one positive scale $\gamma_j$ for the whole filter times a vector of integer codes $\mathbf{p}_j$ from a code set $\mathcal{P}$. We choose between two code sets, ternary $(\mathcal{P}=\{0,\pm1\})$ and powers of two $(\mathcal{P}=\{0,\pm1,\pm2,\pm4\})$. The scale $\gamma_j$ adds no run-time multiply: requantization uses only a right shift, and the remaining scale is folded into the activation's lookup table (Section~\ref{sec:lut}). With the scale folded away, the convolution itself needs no multiplications: a nonzero code is one addition or subtraction of a shifted input ($p\cdot x=\pm(x\ll k)$), and a zero costs nothing.

The share of zeroed weights is chosen to be either fixed, at 70, 50 or 30\%, or free, in which case each weight simply takes the code $p\in\mathcal{P}$ with the lowest error, which may be zero. Section~\ref{sec:projection-sets} compares these sets.

\textbf{Choosing the codes.} Naively rounding each weight to its nearest code minimizes the weight error $\|\mathbf{w}_j-\gamma_j\mathbf{p}_j\|_2^2$, treating all weights alike. However, the filter's output is what actually matters: because its inputs are correlated, errors on different weights can cancel or add up. Data-aware projection therefore minimizes the output error directly, as in post-training quantization~\cite{adaround,obq,gptq}. Let $\mathbf{x}$ be the input samples under the filter's kernel (its \emph{taps}), collected from training windows, and let $\boldsymbol{\mu}=\mathbb{E}[\mathbf{x}]$ be their mean. A weight error $\mathbf{e}_j=\mathbf{w}_j-\gamma_j\mathbf{p}_j$ changes the filter's output by $\mathbf{e}_j^\top\mathbf{x}$, and the variance of that change is $\mathbf{e}_j^\top H\mathbf{e}_j$, where $H$ is the covariance of the taps:
\begin{equation}
 \min_{\gamma_j>0,\,\mathbf{p}_j\in\mathcal{P}} \mathbf{e}_j^\top H\mathbf{e}_j,\qquad H=\mathbb{E}[(\mathbf{x}-\boldsymbol{\mu})(\mathbf{x}-\boldsymbol{\mu})^\top].
 \label{eq:projection}
\end{equation}
To minimize this output error, we solve for the scale and the codes in turn, since each depends on the other, repeating two steps until they stop changing. (1)~With the scale $\gamma_j$ fixed, assign each weight in turn the code in $\mathcal{P}$ that minimizes $\mathbf{e}_j^\top H\mathbf{e}_j$. (2)~With the codes $\mathbf{p}_j$ fixed, set the scale to its least-squares solution $\gamma_j=(\mathbf{p}_j^\top H\mathbf{w}_j)/(\mathbf{p}_j^\top H\mathbf{p}_j)$. We run this from multiple initial scales and retain the solution with the lowest projection error.

With a fixed share of zeroed weights, we search greedily: the largest weights start as their signs ($\pm1$) and the rest as zero. Each round applies the single change (moving a nonzero to a zero position or flipping a sign) that most lowers $\mathbf{e}_j^\top H\mathbf{e}_j$, then refits $\gamma_j$, until no change lowers the error. This gives the final ternary codes; for powers of two, steps (1) and (2) then choose the values at the nonzero positions.

Eq.~\eqref{eq:projection} controls only the variance of the output change, but projection also shifts the filter's mean output by $-\mathbf{e}_j^\top\boldsymbol{\mu}$, which moves $z$ relative to the activation's fixed bounds (Eq.~\eqref{eq:bern}) and can push it out of range. The bias is corrected to cancel this~\cite{dfq}:
\begin{equation}
 b'_j=b_j+\boldsymbol{\mu}^\top\mathbf{e}_j.
\end{equation}

\subsubsection{\textbf{Recovery through learnable Bernstein activations}}
Projection lowers the model's accuracy, and conventional methods recover it by retraining the weights~\cite{twn,inq}. \framework{} instead freezes the projected codes and trains the Bernstein coefficients, along with the convolutional biases, per-filter scales and classifier, in one training run. Because the coefficients directly control each activation's shape (Eq.~\eqref{eq:bern}), training them reshapes each filter's response without changing its frozen weights, which recovers the accuracy lost to projection. Section~\ref{sec:recovery-ablation} shows how much of the recovery the polynomial coefficients carry alone.

Recovery trains on the deployed int8 model through fake-quantized forward passes with straight-through gradients (Section~\ref{sec:lut}). It uses self-distillation with the loss of Eq.~\eqref{eq:kd}, with the unprojected BeCNN as the teacher, and keeps the epoch with the best validation accuracy.

The result is a network with multiplication-free convolutions: only the classifier stays int8 and multiplies. Section~\ref{sec:kernels} generates a kernel that exploits these weights for MCU deployment, and Section~\ref{sec:exp-projection} gives the ReLU baseline, the ablations and the results.

%% file: sections/10_kernels.tex
\subsection{Projection-Aware Kernel Generation and Inference}
\label{sec:kernels}
The last step of \framework{} (Fig.~\ref{fig:framework}) turns the trained BeCNN into integer code for the MCU. The \emph{projection-aware generated kernel} (the \emph{generated kernel} for short) runs projected models (\framework{}$_\text{projected}$), the \emph{dense kernel} runs unprojected ones (\framework{}$_\text{dense}$), and every Bernstein activation becomes a lookup table.

\subsubsection{\textbf{Kernel generation}}
\label{sec:gen-kernel}
\begin{figure*}[t]\centering\includegraphics[width=\textwidth]{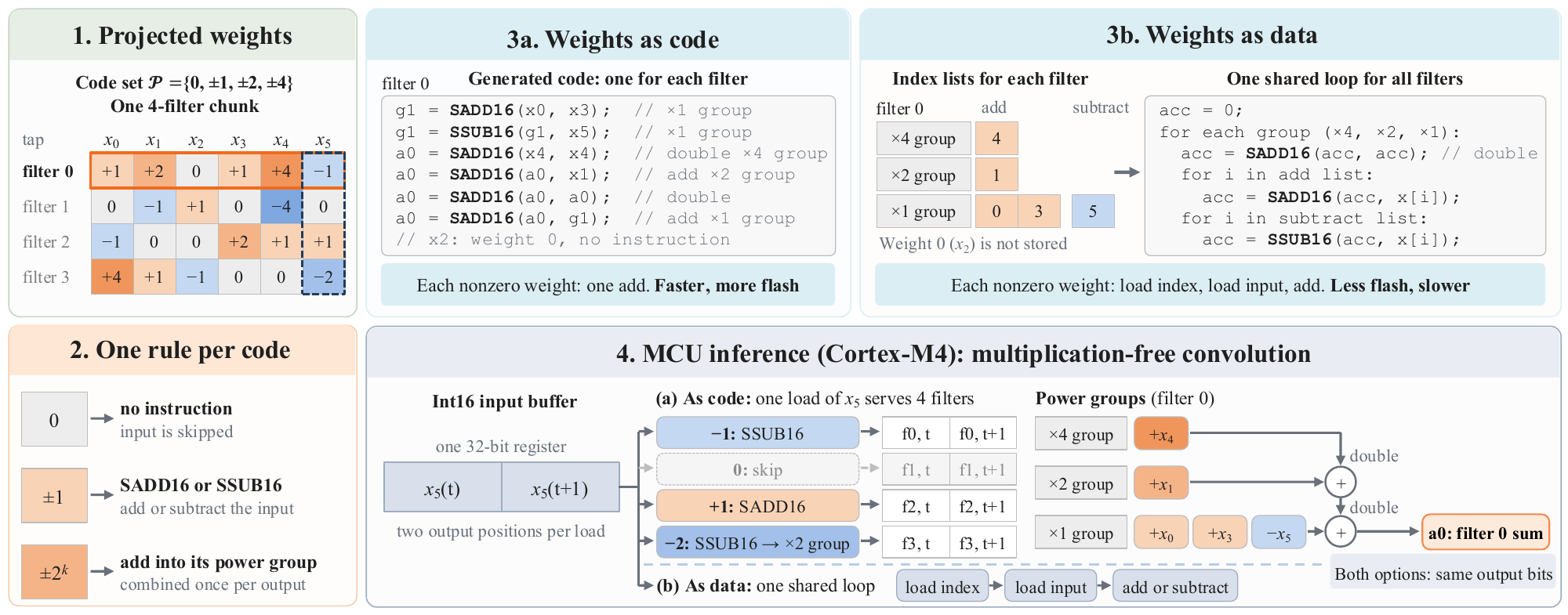}
\Description{A grid of projected weights for four filters and six taps. Each weight value follows one rule: zero emits no instruction, plus or minus one is a SIMD add or subtract, plus or minus a power of two adds into its power group. Filter 0 is shown two ways: as generated code, and as index lists read by one shared loop. On the MCU, one 32-bit register holds an input at two output positions. As code, one load of this input serves four filters; as data, each nonzero weight loads its index, loads its input and adds or subtracts it. Both options sum the same power groups.}
\caption{A schematic four-filter chunk of projected weights, stored two ways with identical outputs. As code (panel 3a), each nonzero weight becomes an addition or subtraction: faster, more flash. As data (panel 3b), index lists drive one shared loop: slower, less flash. Panel 4 shows the Cortex-M4 optimizations.}\label{fig:genkernel}\end{figure*}

Projection alone does not make inference cheaper: a generic kernel still loads and multiplies every weight, zeros included. The savings come from how the kernel stores and applies the fixed codes. \framework{} supports three options: weights as code for speed, weights as data for flash savings, and a mix of the two (Fig.~\ref{fig:genkernel}).

\textbf{Weights as code.} The codes are known at compile time, so a generator writes them directly into the program (Fig.~\ref{fig:genkernel}, panel 3a). Each weight multiplies one tap, so the generator applies one rule per code in $\mathcal{P}$ (panel 2): a zero emits no instruction, $\pm1$ adds or subtracts the input, and $\pm2^k$ adds or subtracts it shifted by $k$. Each filter becomes straight-line code with no weight loads and no loop over weights. This is the fastest option, at the cost of more flash, since every nonzero code becomes instructions.

\textbf{Weights as data.} To save flash, the nonzero codes can instead be stored as index lists read by one shared loop (Fig.~\ref{fig:genkernel}, panel 3b). Each power of two has an add list and a subtract list: for every nonzero code, the loop loads its index, loads that input and adds or subtracts it, doubling the sum between powers. Zeros take no space and all filters share one loop, so this option needs the least flash, at the cost of higher latency from one extra index load per code. It suits large models where flash is the limit.

\textbf{Mix.} The mix sits between these two ends of the trade-off: weights are stored as code for the parts of the model that save the most time per byte, and the rest are stored as data. Section~\ref{sec:exp-storage} compares the three options.

\textbf{Architecture-aware.} The generated kernel also exploits the Cortex-M4's architecture in three ways.
(1)~\emph{One register, two positions.} An int16 input buffer places the inputs of two consecutive output positions in the two halves of one 32-bit register, so each \texttt{SADD16} or \texttt{SSUB16} updates both.
(2)~\emph{Reuse across filters.} With weights as code, one input load serves a chunk of up to four filters, instead of a separate load for each.
(3)~\emph{Power groups.} \texttt{SADD16} has no built-in shift, so shifting every input would cost extra instructions. Instead, each filter adds its inputs into one accumulator per power of two, then combines them once per output by doubling and adding (Fig.~\ref{fig:genkernel}, panel 4). For example, weights $(1,2,-4,1)$ on inputs $(x_0,x_1,x_2,x_3)$ give $g_1=x_0+x_3$, $g_2=x_1$ and $g_4=-x_2$, and output $g_1+2(g_2+2g_4)$.

\subsubsection{\textbf{Lookup-table activations}}
\label{sec:lut}
The kernel's sum $A_j$ then passes through the filter's Bernstein activation. A polynomial is too costly to evaluate at run time on an MCU, but a Bernstein curve's input lies in the fixed range $[0,1]$ (Eq.~\eqref{eq:bern}), so it can be precomputed into a lookup table. Bern2Edge~\cite{bern2edge} introduced such tables for FPGAs, with linear interpolation that costs one multiply per activation. \framework{} removes this multiply: each input maps directly to one table entry, so an activation is a single table load.

\textbf{Table indexing.} Each filter's curve is stored in a table $\mathrm{LUT}_j$ of $E=64$ int8 entries, and finding an entry at run time takes only an addition, a shift and a clamp. The normalization of Eq.~\eqref{eq:bern} is computed ahead of time: its division and $l_j$ fold into the filter's scale $s_j$ and integer bias $\bar b_j$, so $t=s_j(A_j+\bar b_j)$, where $s_j$ includes $\gamma_j$ for a projected model. A right shift by $r_j$, chosen per filter at compile time as the smallest shift for which the $E$ entries cover all of $[0,1]$, brings this sum down to a table index:
\begin{equation}
 q_j=\operatorname{sat}\big((A_j+\bar b_j)\mathbin{\gg}r_j\big),\qquad
 \hat\sigma_j=\mathrm{LUT}_j[q_j],
 \label{eq:lut}
\end{equation}
where $\operatorname{sat}$ clamps to $[0,\,E-1]$, keeping $t$ in $[0,1]$ as in Eq.~\eqref{eq:bern}.

\textbf{Scale-aware tables.} A shift divides only by powers of two, so it cannot match the filter's scale $s_j$ exactly. Correcting this at run time would take a multiply, so each table instead stores the curve at the inputs its entries actually receive, adapting scale-aware activation tables and shift-based scaling~\cite{tflitelut,bitshiftquant}. The remaining constants also need no run-time work: the table's output scale and offset fold into the classifier, and BatchNorm and the pooling average fold into adjacent layers.

\textbf{Same cost at any degree.} A curve of any degree compiles to the same $E$-entry table, so the degree changes only the stored values, not the flash ($E$ bytes per curve), RAM or instructions (one load per output). High-degree polynomials therefore add expressiveness under compression and capacity for recovery at no additional deployed inference cost.

The resulting convolution needs no multiplications: weights become additions, requantization a shift and the activation a table read. Only the int8 classifier still multiplies.

\subsubsection{\textbf{Dense kernel}}
\label{sec:dense-kernel}
The dense kernel runs the unprojected BeCNNs. It stores weights as data and multiplies each one, with three optimizations. (1)~\emph{SIMD multiply-accumulates.} \texttt{SMLAD} performs two per instruction, as in CMSIS-NN~\cite{cmsisnn}, and one pass computes two time positions of four filters in eight register accumulators. (2)~\emph{Direct inputs.} The input is widened once into a zero-padded int16 buffer, so filters read it in place instead of copying it per position. (3)~\emph{Fusion.} Each sum goes straight through the shift, the activation table of Eq.~\eqref{eq:lut} and the pooling sum, without being stored. Section~\ref{sec:exp-kernels} compares this kernel with TFLM+CMSIS-NN and with the generated kernel.

%% file: sections/11_extensions.tex
\subsection{Extensions to Other Architectures}
\label{sec:extensions}
\framework{} is not specific to convolutions: any linear layer followed by a scalar activation can use the same pipeline. In a transformer, per-unit Bernstein curves replace the feed-forward network's activation (GELU or Swish), and projection makes the linear layers multiplication-free; only the attention products, which multiply two activations, keep their multiplications. In an MLP, each hidden neuron gets one curve, and the hidden rows are projected to multiplication-free codes. Section~\ref{sec:exp-extensions} evaluates both.

%% file: sections/12_results_setup.tex
\section{Results}
\subsection{Experimental Setup}
\label{sec:setup}
\paragraph{\textbf{Terminology.}} Throughout, BeCNN denotes the model architecture, a CNN with learnable Bernstein activations, and \framework{} the complete optimization and deployment framework. We call the trained BeCNN before weight projection \framework{}$_\text{dense}$, and the recovered model with frozen multiplication-free weight codes \framework{}$_\text{projected}$.

\paragraph{\textbf{Datasets.}} Table~\ref{tab:data} lists the tasks: three HAR datasets~\cite{ucihar,hhar,pamap}, split by person; Speech Commands v2~\cite{speechcommands}, with the MLPerf Tiny front end~\cite{mlperftiny} for the depthwise-separable CNNs (DS-CNN) and all 35 words for the keyword transformers; and UJIIndoorLoc~\cite{torressospedra2014uji} for the MLPs. Appendix~\ref{app:setup-details} gives the windowing and preprocessing.

\begin{table}[t]
\caption{Tasks and splits.}
\label{tab:data}
\centering\small\setlength{\tabcolsep}{2pt}
\begin{tabular*}{\columnwidth}{@{\extracolsep{\fill}}llcrrr@{}}\toprule
Task & Input & Classes & Train & Val & Test\\\midrule
UCI-HAR & $6\times128$ & 6 & 5,867 & 1,485 & 2,947\\
HHAR & $6\times128$ & 6 & 30,850 & 5,374 & 10,755\\
PAMAP2 & $18\times128$ & 12 & 10,982 & 2,107 & 1,934\\
Speech Cmds v2 & $49\times10$ MFCC & 12 & 85,511 & 10,102 & 4,890\\
Speech Cmds v2 (KWT) & $98\times40$ MFCC & 35 & 84,843 & 9,981 & 11,005\\
UJIIndoorLoc & 520 RSSI & 3+5  & 13,955 & 5,982 & 1,111\\
\bottomrule\end{tabular*}
\end{table}

\paragraph{\textbf{Teacher models.}} Each task has one teacher, shared by its Bernstein and ReLU students. ResNet~\cite{wang2017resnet} (UCI-HAR), DeepSense~\cite{yao2017deepsense} (HHAR) and DeepConvLSTM~\cite{ordonez2016deepconvlstm} (PAMAP2) are published architectures trained with their published optimizers and settings. HAR students distill from the logit average of three seeds, while reported teacher accuracies are per seed. The DS-CNN teacher is the released MLPerf Tiny reference model~\cite{mlperftiny,mlperfkws}. The keyword transformers are distilled from AST~\cite{gong2021ast} and the UJIIndoorLoc MLPs from the MLP of Kim et al.~\cite{kim2018indoor} (Section~\ref{sec:exp-extensions}). Appendix~\ref{app:setup-details} lists all teachers with their accuracies and training recipes.

\paragraph{\textbf{Training hardware.}} All models are trained in PyTorch on NVIDIA A30, V100 and A100 GPUs. Against a ReLU student of the same shape, a Bernstein student's curve coefficients add 20--178\% to the convolution's parameters, and an epoch takes $1.32\times$ as long. Inference does not pay this cost, because each curve compiles into a fixed lookup table (Section~\ref{sec:lut}).

\paragraph{\textbf{Board and QEMU}} The target is the TI CC3220SF LaunchPad: an 80\,MHz Cortex-M4 with 256\,KiB RAM, 1\,MiB flash, no floating-point unit and no flash cache~\cite{cc3220sf,cc3220trm}; code runs from flash. Instruction counts come from QEMU's Cortex-M4 machine with the same build flags (Appendix~\ref{app:setup-details}): the inference machine code is identical to the board's, and a single-step count of one inference on the board equals QEMU's. Every image is checked bit for bit against the integer reference.

\paragraph{\textbf{Metrics.}} Accuracy is the test accuracy of the int8 network, whose outputs equal the MCU's, as mean $\pm$ population SD over three seeds; accuracy differences are in percentage points (pp). Instructions are QEMU instructions for one inference. Latency is measured on the CC3220SF board for the seed-0 model: Cortex-M4 cycle-counter cycles around one inference, median over at least 100 test inputs, at 80\,MHz. Flash counts the model and runtime without the test harness; RAM counts static data and peak stack.

\paragraph{\textbf{DS-CNN activations.}} A DS-CNN has an activation after every layer, so one curve per filter would cost far more flash than on the HAR BeCNNs. Its BeCNN students therefore share one Bernstein curve per layer, and BatchNorm gives each filter its own input scale and shift on that curve.

%% file: sections/13_results_pipeline.tex
\subsection{\framework{} End-to-End Results}
\label{sec:exp-pipeline}

Table~\ref{tab:e2e} presents \framework{}'s end-to-end results, before (\framework{}$_\text{dense}$) and after (\framework{}$_\text{projected}$) projection, against the teacher baseline across all tasks. The teachers run on TensorFlow Lite Micro (TFLM)~\cite{tflm} with Arm's CMSIS-NN kernels~\cite{cmsisnn}. TFLM is also the runtime of MLPerf Tiny's reference implementations~\cite{mlperftiny}, making this a strong, widely reproduced baseline. Appendix~\ref{app:tflm_base} compares TFLM's plain C reference kernels, which run 2.5 to 6.1$\times$ slower.

\textbf{Accuracy.} Across the six tasks, the projected students lose 0.7\,pp on average against their int8 teachers and 1.56\,pp against their float32 teachers. They beat the int8 teachers on PAMAP2 and UJIIndoorLoc.

\textbf{Hardware.} The HAR teachers are not built for microcontrollers, so the large reductions come mostly from compression. On UCI-HAR and HHAR, the projected students run 834 to 9,580$\times$ faster than the teachers, with 36 to 71$\times$ less flash and about 20 to 31$\times$ less RAM. DeepConvLSTM does not fit the board at all (a 290\,KiB tensor arena against 256\,KiB of RAM). AST does not fit either. These teacher-to-student ratios therefore measure the whole framework, dominated by compression, rather than the kernels alone. The same-architecture and dense-versus-projected comparisons of Sections~\ref{sec:exp-kd}--\ref{sec:exp-kernels} isolate the contributions of the Bernstein activations, projection and generated kernel.

The MLPerf Tiny DS-CNN, in contrast, is a strong model built for microcontrollers, and \framework{} still beats it. \framework{}$_\text{dense}$ is 1.0\,pp more accurate and $2.2\times$ faster, with $6.4\times$ less flash and $2.0\times$ less RAM. \framework{}$_\text{projected}$ gives up 1.0\,pp for $5.9\times$ lower latency, $2.3\times$ less flash and $1.5\times$ less RAM. Its generated kernel trades flash for speed, turning the weights into instructions (Section~\ref{sec:kernels}). 

Against the same model on TinyEngine~\cite{mcunet}, a code generator with no interpreter, \framework{}$_\text{dense}$ wins on accuracy, latency, flash and RAM, while \framework{}$_\text{projected}$ is $6.6\times$ faster but needs $1.5\times$ more flash and loses 1.0\,pp of accuracy. TinyEngine's outputs match TFLM's, and it uses $3.5\times$ less flash and $1.4\times$ less RAM, but it is 13\% slower. TinyEngine has no kernel for the model's $10{\times}4$ first convolution and no int8 softmax, so these two layers use the CMSIS-NN kernels bundled with it, included in its latency and memory. TinyEngine also cannot run the HAR teachers: it has no kernels for their $1{\times}k$ convolutions and no recurrent layers.

\begin{table}[t]
\caption{Each task's int8 teacher on TFLM with CMSIS-NN against \framework{}. Parentheses: teacher's float32 accuracy; ($\downarrow\times$): TFLM teacher / \framework{}. UJIIndoorLoc: joint building-and-floor accuracy. $^\ast$Does not fit the board: AST measured in QEMU on TFLM's reference kernels, so it has no latency; DeepConvLSTM has no Cortex-M build, so its values are lower bounds.}
\label{tab:e2e}
\centering\footnotesize\setlength{\tabcolsep}{0.7pt}\renewcommand{\arraystretch}{1.1}
\begin{tabular*}{\columnwidth}{@{\extracolsep{\fill}}llrrrr@{}}
\toprule
Task & Model & Acc.\ (\%) & Lat. (ms) & Flash (KiB) & RAM (KiB)\\
\midrule
\multirow{3}{*}{UCI-HAR} & ResNet & 96.01\,$\pm$\,0.32 (96.52) & 4,569 & 610.9 & 77.6\\
\cmidrule(l){2-6}
 & \framework{}$_\text{dense}$ & 94.70\,$\pm$\,0.09 & \rd{0.92}{4,971} & \rd{11.9}{51.2} & \rd{2.5}{31.6}\\
 & \framework{}$_\text{projected}$ & 93.96\,$\pm$\,0.79 & \rd{0.48}{9,580} & \rd{8.6}{70.7} & \rd{2.5}{31.4}\\
\midrule
\multirow{3}{*}{HHAR} & DeepSense & 94.72\,$\pm$\,0.57 (95.08) & 934.4 & 660.1 & 52.9\\
\cmidrule(l){2-6}
 & \framework{}$_\text{dense}$ & 92.91\,$\pm$\,0.25 & \rd{2.49}{375} & \rd{16.6}{39.8} & \rd{2.7}{19.3}\\
 & \framework{}$_\text{projected}$ & 92.16\,$\pm$\,0.53 & \rd{1.12}{834} & \rd{18.1}{36.4} & \rd{2.7}{19.7}\\
\midrule
\multirow{3}{*}{PAMAP2} & DeepConvLSTM$^\ast$ & 84.13\,$\pm$\,1.86 (86.99) & --- & $\ge$849.5 & $>$256.0\\
\cmidrule(l){2-6}
 & \framework{}$_\text{dense}$ & 86.92\,$\pm$\,0.97 & 1.54 & \rd{13.0}{$\ge$65.5} & \rd{5.5}{$>$46.4}\\
 & \framework{}$_\text{projected}$ & 86.73\,$\pm$\,0.86 & 0.79 & \rd{9.6}{$\ge$88.4} & \rd{5.9}{$>$43.5}\\
\midrule
\multirow{4}{*}{KWS} & MLPerf Tiny & 91.62 (92.17) & 259.6 & 134.8 & 24.1\\
 & On TinyEngine & 91.62 & 294.1 & 38.3 & 16.8\\
\cmidrule(l){2-6}
 & \framework{}$_\text{dense}$ & 92.60\,$\pm$\,0.86 & \rd{120.2}{2.16} & \rd{20.9}{6.44} & \rd{12.0}{2.00}\\
 & \framework{}$_\text{projected}$ & 90.62\,$\pm$\,1.01 & \rd{44.3}{5.86} & \rd{58.3}{2.31} & \rd{15.9}{1.51}\\
\midrule
\multirow{3}{*}{KWT} & AST$^\ast$ & 97.10 (98.03) & --- & 83,797.0 & 2,353.9\\
\cmidrule(l){2-6}
 & \framework{}$_\text{dense}$ & 95.68\,$\pm$\,0.11 & 1,859 & \rd{249.5}{336} & \rd{60.5}{38.9}\\
 & \framework{}$_\text{projected}$ & 95.44\,$\pm$\,0.21 & 1,529 & \rd{246.3}{340} & \rd{62.5}{37.7}\\
\midrule
\multirow{3}{*}{UJIIndoorLoc} & MLP & 92.02\,$\pm$\,0.64 (92.02) & 14.7 & 211.4 & 2.88\\
\cmidrule(l){2-6}
 & \framework{}$_\text{dense}$ & 92.59\,$\pm$\,0.36 & \rd{3.34}{4.40} & \rd{91.9}{2.30} & \rd{2.41}{1.19}\\
 & \framework{}$_\text{projected}$ & 92.56\,$\pm$\,0.04 & \rd{0.60}{24.4} & \rd{65.4}{3.23} & \rd{1.20}{2.41}\\
\bottomrule
\end{tabular*}
\end{table}

\textbf{Without a hardware multiplier.} Table~\ref{tab:e2e-nomul} repeats the comparison on RV32EC~\cite{riscvspec}, a RISC-V instruction set with no multiply instruction. The teachers therefore spend 83 to 93\% of their instructions in a software routine, against 13 to 29\% for \framework{}$_\text{projected}$. On the MLPerf Tiny reference, the reduction in instructions grows from 5.2$\times$ on the Cortex-M4 to 147$\times$. As a control, the same RISC-V core with the multiply extension (RV32EMC) reaches only 7.9$\times$ (Appendix~\ref{app:rv32emc}), so the larger reduction comes from the missing multiplier, not from RISC-V.

\begin{table}[t]
\caption{Teachers against \framework{}$_\text{projected}$: QEMU instructions per inference on the Cortex-M4 and RV32EC, with RV32EC flash and RAM; ($\downarrow\times$): TFLM teacher / student. ---: no Cortex-M build. On RV32EC, teachers use CMSIS-NN's portable C code, with TinyEngine's Arm DSP instructions rewritten in plain C.}
\label{tab:e2e-nomul}
\centering\footnotesize\setlength{\tabcolsep}{1pt}
\renewcommand{\arraystretch}{1.3}
\begin{tabular*}{\columnwidth}{@{\extracolsep{\fill}}llrrrrr@{}}
\toprule
 & & & \multicolumn{2}{c}{Instr. (M)} & Flash & RAM\\
\cmidrule(lr){4-5}
Task & Model & Acc.\ (\%) & M4 & RV32EC & (KiB) & (KiB)\\
\midrule
\multirow{2}{*}{UCI-HAR} & ResNet & 96.01\,$\pm$\,0.32 & 120.4 & 5,103.6 & 621.6 & 77.4\\
 & \framework{}$_\text{projected}$ & 93.96\,$\pm$\,0.79 & \rd{0.013}{8,962} & \rd{0.093}{54,798} & \rd{25.8}{24.1} & \rd{3.2}{24.4}\\
\midrule
\multirow{2}{*}{HHAR} & DeepSense & 94.72\,$\pm$\,0.57 & 24.8 & 790.7 & 682.9 & 53.3\\
 & \framework{}$_\text{projected}$ & 92.16\,$\pm$\,0.53 & \rd{0.032}{767} & \rd{0.274}{2,890} & \rd{82.0}{8.3} & \rd{3.5}{15.4}\\
\midrule
\multirow{2}{*}{PAMAP2} & DeepConvLSTM & 84.13\,$\pm$\,1.86 & --- & 25,770.9 & 941.5 & 291.8\\
 & \framework{}$_\text{projected}$ & 86.73\,$\pm$\,0.86 & 0.022 & \rd{0.167}{154,680} & \rd{48.7}{19.3} & \rd{8.2}{35.4}\\
\midrule
\multirow{3}{*}{KWS} & MLPerf Tiny & 91.62 & 7.6 & 439.3 & 148.3 & 23.9\\
 & On TinyEngine & 91.62 & 8.6 & 251.5 & 42.8 & 16.5\\
 & \framework{}$_\text{projected}$ & 90.62\,$\pm$\,1.01 & \rd{1.460}{5.2} & \rd{2.996}{147} & \rd{88.5}{1.7} & \rd{18.5}{1.3}\\
\bottomrule
\end{tabular*}
\end{table}

%% file: sections/14_results_nas.tex
\subsection{Hardware-Aware Neural Architecture Search}
\label{sec:exp-nas}\label{sec:exp-nas-setup}\label{sec:exp-nas-kws}
\textbf{Setup.} The search of Section~\ref{sec:nas} runs on two model families. HAR CNNs use the search space of Table~\ref{tab:nas-space}. DS-CNNs for keyword spotting have at most five convolution layers, and the search covers the width (16--128), the first convolution's kernel and stride, the curves per channel at the last activation, the learning rate and, for Bernstein, the degree (15, 21 or 31). Trials train for 60 epochs on labels only, are pruned at epochs 7 and 20, and are selected on validation accuracy. Bernstein studies receive more trials than ReLU studies (500 vs.\ 300 for HAR, 200 vs.\ 150 for DS-CNNs): their extra dimensions (degree and the coefficient learning-rate multiplier) change neither the compiled kernel nor its instructions, flash or RAM, so at equal trial counts Bernstein would cover fewer distinct architectures. Every architecture uses at most 16\% of the CC3220SF's flash and 13\% of its RAM, so the memory constraints of Eq.~\eqref{eq:nas} never bind and instructions alone shape the fronts (Appendix~\ref{app:nas}). Instructions rank models as board cycles do (Spearman 0.9964 over 180 board images).

\begin{table}[t]\caption{HAR search space. The ReLU control omits the last two rows.}\label{tab:nas-space}
\centering\small\setlength{\tabcolsep}{2pt}
\begin{tabular*}{\columnwidth}{@{\extracolsep{\fill}}ll@{}}\toprule Dimension & Values\\\midrule
Dilation groups $G$ & 1, 2, 3 (dilations $1,\dots,2^{G-1}$)\\
Filters per group $f$ & 8, 12, 16, 24, 32, 48, 64, 96, 128\\
Kernel size $k$ & 3, 5, 7, 9\\
Stride $s$ (all groups) & 1, 2, 4, 8, 16, 32\\
Normalization & none, BatchNorm\\
Learning rate & $3\times10^{-3}$, $10^{-2}$, $3\times10^{-2}$\\
Bernstein degree $N$ & 3, 5, 7, 11, 15, 23, 31\\
Coefficient lr multiplier & 1, 3, 10\\\bottomrule\end{tabular*}\end{table}

\begin{figure*}[t]\centering\includegraphics[width=\textwidth]{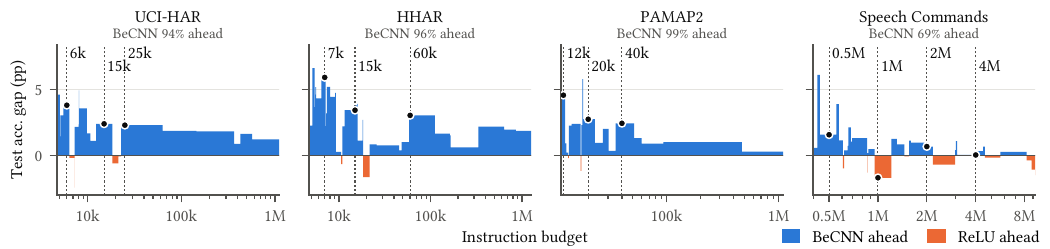}
\Description{Four panels, one per task, show the test accuracy difference between the BeCNN and the ReLU model selected under each instruction budget, on a log axis. Blue areas above zero mark budgets where the BeCNN is ahead and orange areas below zero where ReLU is ahead. Blue covers almost all of the range on UCI-HAR, HHAR and PAMAP2 and most of the range below one million instructions on Speech Commands. Dashed lines mark the budgets used in the paper.}
\caption{Int8 test accuracy of the BeCNN minus the ReLU model, in pp, for the models selected on validation accuracy at every instruction budget. Dashed lines: the budgets carried forward.}
\label{fig:nas-budget-gap}\end{figure*}

\textbf{Results.} Figure~\ref{fig:nas-budget-gap} applies the budget rule of Section~\ref{sec:nas}, which selects on the three-seed validation mean, at every instruction budget and plots the test-accuracy gap between the two picks. On HAR, the BeCNN is almost always ahead, over 94 to 99\% of the log-instruction range. On Speech Commands, the BeCNN leads over 69\% of the range, mostly below 1M, and in the intermediate budgets the two trade the lead within approximately 1\,pp.

These gaps come from cheap training, so they rank architectures rather than give final accuracy. We carry three budgets per HAR dataset and four for Speech Commands through the rest of the \framework{} pipeline (Tables~\ref{tab:nas-budget} and~\ref{tab:nas-budget-kws}). On HAR, the BeCNN leads at all budgets by 2.31 to 5.94\,pp, with its largest lead at each dataset's smallest budget. On Speech Commands it leads at 0.5M and 2M, ReLU leads at 1M, and the two tie at 4M. Full training and projection change these gaps (Sections~\ref{sec:exp-kd} and~\ref{sec:exp-projection}).

\begin{table}[t]
\caption{Best model per instruction budget (HAR). Shape: dilation groups$\times$filters per group, kernel size (k), stride (s), BatchNorm (BN), Bernstein degree (N). $\Delta$Acc.: BeCNN minus ReLU, in pp. Models are selected on validation accuracy; Test: int8 test accuracy, three seeds.}
\label{tab:nas-budget}
\centering\small\setlength{\tabcolsep}{2pt}\setlength{\aboverulesep}{0pt}\setlength{\belowrulesep}{0pt}\renewcommand{\arraystretch}{1.15}
\begin{tabular*}{\columnwidth}{@{\extracolsep{\fill}}c|l|lll|cc@{}}
\toprule
Task & Budget & Model & Shape & Instr. & Test (\%) & $\Delta$Acc.\\\midrule
\multirow{6}{*}{\rotatebox[origin=c]{90}{UCI-HAR}}
 & \multirow{2}{*}{6k}  & BeCNN & 1$\times$8, k7, s32, N5 & 5,530 & $\mathbf{83.33}\pm0.63$ & \multirow{2}{*}{+3.82}\\
 &                      & ReLU & 1$\times$8, k7, s32 & 5,497 & $79.50\pm0.07$ & \\\cmidrule{2-7}
 & \multirow{2}{*}{15k} & BeCNN & 1$\times$8, k9, s8, N31 & 12,334 & $\mathbf{91.81}\pm1.01$ & \multirow{2}{*}{+2.41}\\
 &                      & ReLU & 3$\times$8, k3, s16 & 10,727 & $89.40\pm0.65$ & \\\cmidrule{2-7}
 & \multirow{2}{*}{25k} & BeCNN & 3$\times$12, k3, s8, BN, N31 & 22,841 & $\mathbf{94.23}\pm0.29$ & \multirow{2}{*}{+2.31}\\
 &                      & ReLU & 3$\times$16, k5, s16 & 23,189 & $91.92\pm0.15$ & \\\midrule
\multirow{6}{*}{\rotatebox[origin=c]{90}{HHAR}}
 & \multirow{2}{*}{7k}  & BeCNN & 1$\times$8, k5, s16, BN, N23 & 6,438 & $\mathbf{81.36}\pm1.12$ & \multirow{2}{*}{+5.94}\\
 &                      & ReLU & 1$\times$8, k5, s16, BN & 6,405 & $75.42\pm2.26$ & \\\cmidrule{2-7}
 & \multirow{2}{*}{15k} & BeCNN & 2$\times$8, k7, s16, BN, N23 & 11,536 & $\mathbf{86.71}\pm1.23$ & \multirow{2}{*}{+3.44}\\
 &                      & ReLU & 3$\times$8, k7, s32, BN & 10,547 & $83.27\pm0.90$ & \\\cmidrule{2-7}
 & \multirow{2}{*}{60k} & BeCNN & 3$\times$16, k9, s8, N11 & 58,688 & $\mathbf{91.75}\pm0.68$ & \multirow{2}{*}{+3.05}\\
 &                      & ReLU & 3$\times$24, k9, s16, BN & 48,371 & $88.69\pm0.25$ & \\\midrule
\multirow{6}{*}{\rotatebox[origin=c]{90}{PAMAP2}}
 & \multirow{2}{*}{12k} & BeCNN & 1$\times$8, k3, s32, BN, N11 & 11,424 & $\mathbf{79.15}\pm1.22$ & \multirow{2}{*}{+4.58}\\
 &                      & ReLU & 1$\times$8, k3, s32 & 11,321 & $74.56\pm1.83$ & \\\cmidrule{2-7}
 & \multirow{2}{*}{20k} & BeCNN & 1$\times$8, k3, s8, BN, N11 & 17,352 & $\mathbf{85.09}\pm1.49$ & \multirow{2}{*}{+2.76}\\
 &                      & ReLU & 1$\times$16, k3, s16 & 18,035 & $82.33\pm0.90$ & \\\cmidrule{2-7}
 & \multirow{2}{*}{40k} & BeCNN & 1$\times$16, k5, s8, BN, N31 & 35,112 & $\mathbf{87.18}\pm1.88$ & \multirow{2}{*}{+2.45}\\
 &                      & ReLU & 1$\times$8, k9, s8, BN & 30,449 & $84.73\pm0.51$ & \\
\bottomrule\end{tabular*}
\end{table}

\begin{table}[t]
\caption{Best DS-CNN per instruction budget. Shape: width (w), first-convolution kernel (k$\times$4) and stride (s$\times$2), curves per channel at the last activation (h), Bernstein degree (N). Models are selected on validation accuracy; Test: int8 test accuracy, three seeds.}
\label{tab:nas-budget-kws}
\centering\small\setlength{\tabcolsep}{2pt}\setlength{\aboverulesep}{0pt}\setlength{\belowrulesep}{0pt}\renewcommand{\arraystretch}{1.15}
\begin{tabular*}{\columnwidth}{@{\extracolsep{\fill}}l|llr|cc@{}}
\toprule
Budget & Model & Shape & Instr. & Test (\%) & $\Delta$Acc.\\\midrule
\multirow{2}{*}{0.5M} & BeCNN & w16, k6, s4, h4, N15 & 423,591 & $\mathbf{79.14}\pm1.46$ & \multirow{2}{*}{+1.58}\\
                      & ReLU & w16, k6, s4, h4      & 438,713 & $77.56\pm1.03$ & \\\midrule
\multirow{2}{*}{1M}   & BeCNN & w32, k6, s4, h1, N15 & 872,678 & $86.00\pm0.11$ & \multirow{2}{*}{$-1.69$}\\
                      & ReLU & w32, k6, s4, h4      & 953,880 & $\mathbf{87.69}\pm0.52$ & \\\midrule
\multirow{2}{*}{2M}   & BeCNN & w48, k6, s3, h1, N15 & 1,934,222 & $\mathbf{89.86}\pm0.58$ & \multirow{2}{*}{+0.68}\\
                      & ReLU & w48, k10, s4, h1     & 1,595,564 & $89.18\pm0.21$ & \\\midrule
\multirow{2}{*}{4M}   & BeCNN & w64, k6, s3, h4, N21 & 3,023,118 & $\mathbf{91.64}\pm0.23$ & \multirow{2}{*}{+0.04}\\
                      & ReLU & w48, k10, s2, h1     & 3,077,451 & $91.60\pm0.57$ & \\
\bottomrule
\end{tabular*}
\end{table}

%% file: sections/15_results_kd.tex
\subsection{BeCNN Training via Knowledge Distillation and Dense Deployment}
\label{sec:exp-kd}

\textbf{Setup.} The budget models of Tables~\ref{tab:nas-budget} and~\ref{tab:nas-budget-kws} are distilled from their task's teacher, with the same recipe for both activations. We use $T=4$, $\alpha \in \{0, 0.5, 0.7\}$ ($\alpha=0$: labels only), AdamW with a cosine learning-rate schedule, and early stopping on validation accuracy. For each model we report the $\alpha$ with the best three-seed accuracy mean.

\textbf{Results.} At the largest budgets, the distilled BeCNNs come within 1.3 and 1.8\,pp, respectively, of the int8 teachers on UCI-HAR and HHAR, and beat them on PAMAP2 and Speech Commands (Tables~\ref{tab:kd-budget} and~\ref{tab:kd-budget-kws}). The BeCNN also beats its ReLU counterpart at all nine HAR budgets, by 0.2 to 4.5\,pp, and at three of four Speech Commands budgets. Its lead is largest at the smallest budget on most tasks, showing that the learned activations help most when a network has little capacity left. Every student fits the CC3220SF, with at most 20.9\,KiB of flash and 12.3\,KiB of RAM.

\textbf{Same-architecture control.} To isolate the effect of the NAS and of KD from that of the Bernstein activation, we also train both activations on the same nine architectures, without KD (Table~\ref{tab:same-shape-board}). The BeCNN leads on eight pairs and ties on the ninth, by 2.75\,pp on average, showing that the advantage comes mainly from the activation itself. Its hardware overhead is small: 0.5 to 4.8\% more latency and at most 3.4\,KiB more flash (0.3\% of the board's 1\,MiB).

\begin{table}[t]
\caption{HAR NAS models (Table~\ref{tab:nas-budget}) after KD. Test: int8, three seeds. Flash and RAM (KiB) and latency (ms) on the CC3220SF. $\Delta$Acc.: BeCNN minus ReLU (pp).}
\label{tab:kd-budget}
\centering\small\setlength{\tabcolsep}{2pt}\setlength{\aboverulesep}{0pt}\setlength{\belowrulesep}{0pt}\renewcommand{\arraystretch}{1.15}
\begin{tabular*}{\columnwidth}{@{\extracolsep{\fill}}c|ll|rr|rrr@{}}
\toprule
Task & Budget & Model & Test (\%) & $\Delta$Acc. & Flash & RAM & Lat.\\\midrule
\multirow{6}{*}{\rotatebox[origin=c]{90}{UCI-HAR}}
 & \multirow{2}{*}{6k}  & BeCNN & $\mathbf{83.92}\pm0.08$ & \multirow{2}{*}{+2.73} & 9.1 & 2.4 & 0.221\\
 &                      & ReLU  & $81.19\pm1.01$ & & 8.2 & 2.4 & 0.215\\\cmidrule{2-8}
 & \multirow{2}{*}{15k} & BeCNN & $\mathbf{92.29}\pm1.12$ & \multirow{2}{*}{+2.32} & 9.3 & 2.5 & 0.511\\
 &                      & ReLU  & $89.97\pm0.61$ & & 8.7 & 2.4 & 0.418\\\cmidrule{2-8}
 & \multirow{2}{*}{25k} & BeCNN & $\mathbf{94.70}\pm0.09$ & \multirow{2}{*}{+1.19} & 11.9 & 2.5 & 0.919\\
 &                      & ReLU  & $93.51\pm0.73$ & & 11.0 & 2.5 & 0.926\\\midrule
\multirow{6}{*}{\rotatebox[origin=c]{90}{HHAR}}
 & \multirow{2}{*}{7k}  & BeCNN & $\mathbf{81.55}\pm0.27$ & \multirow{2}{*}{+4.53} & 8.9 & 2.4 & 0.257\\
 &                      & ReLU  & $77.02\pm0.66$ & & 8.0 & 2.4 & 0.249\\\cmidrule{2-8}
 & \multirow{2}{*}{15k} & BeCNN & $\mathbf{88.68}\pm1.02$ & \multirow{2}{*}{+4.40} & 10.4 & 2.5 & 0.477\\
 &                      & ReLU  & $84.28\pm1.02$ & & 9.9 & 2.6 & 0.426\\\cmidrule{2-8}
 & \multirow{2}{*}{60k} & BeCNN & $\mathbf{92.91}\pm0.25$ & \multirow{2}{*}{+0.21} & 16.6 & 2.7 & 2.491\\
 &                      & ReLU  & $92.70\pm0.13$ & & 16.0 & 2.7 & 1.983\\\midrule
\multirow{6}{*}{\rotatebox[origin=c]{90}{PAMAP2}}
 & \multirow{2}{*}{12k} & BeCNN & $\mathbf{79.99}\pm0.37$ & \multirow{2}{*}{+2.57} & 10.4 & 5.4 & 0.448\\
 &                      & ReLU  & $77.42\pm0.57$ & & 9.5 & 5.4 & 0.446\\\cmidrule{2-8}
 & \multirow{2}{*}{20k} & BeCNN & $\mathbf{85.92}\pm0.75$ & \multirow{2}{*}{+0.74} & 10.4 & 5.4 & 0.715\\
 &                      & ReLU  & $85.18\pm0.30$ & & 10.5 & 5.4 & 0.754\\\cmidrule{2-8}
 & \multirow{2}{*}{40k} & BeCNN & $\mathbf{86.92}\pm0.97$ & \multirow{2}{*}{+2.79} & 13.0 & 5.5 & 1.541\\
 &                      & ReLU  & $84.13\pm0.99$ & & 11.2 & 5.6 & 1.360\\
\bottomrule\end{tabular*}
\end{table}

\begin{table}[t]
\caption{DS-CNN NAS models (Table~\ref{tab:nas-budget-kws}) after distillation; Test: int8, three seeds. Flash and RAM (KiB) and latency (ms) on the CC3220SF. $\Delta$Acc.: BeCNN minus ReLU (pp).}
\label{tab:kd-budget-kws}
\centering\small\setlength{\tabcolsep}{2pt}\setlength{\aboverulesep}{0pt}\setlength{\belowrulesep}{0pt}\renewcommand{\arraystretch}{1.15}
\begin{tabular*}{\columnwidth}{@{\extracolsep{\fill}}ll|rr|rrr@{}}
\toprule
Budget & Model & Test (\%) & $\Delta$Acc. & Flash & RAM & Lat.\\\midrule
\multirow{2}{*}{0.5M} & BeCNN & $\mathbf{81.07}\pm0.35$ & \multirow{2}{*}{+2.09} & 8.0 & 2.7 & 17.4\\
                      & ReLU  & $78.98\pm0.54$ & & 7.8 & 2.7 & 16.9\\\midrule
\multirow{2}{*}{1M}   & BeCNN & $86.41\pm0.20$ & \multirow{2}{*}{$-0.82$} & 10.0 & 4.6 & 35.0\\
                      & ReLU  & $\mathbf{87.23}\pm0.59$ & & 11.3 & 5.0 & 36.5\\\midrule
\multirow{2}{*}{2M}   & BeCNN & $\mathbf{90.74}\pm0.46$ & \multirow{2}{*}{+0.54} & 13.8 & 8.6 & 76.8\\
                      & ReLU  & $90.20\pm1.63$ & & 14.2 & 6.7 & 60.8\\\midrule
\multirow{2}{*}{4M}   & BeCNN & $\mathbf{92.60}\pm0.86$ & \multirow{2}{*}{+0.94} & 20.9 & 12.0 & 120.2\\
                      & ReLU  & $91.66\pm0.63$ & & 14.2 & 12.3 & 117.1\\
\bottomrule\end{tabular*}
\end{table}

\begin{table*}[t]
\caption{BeCNN and ReLU CNN with identical architectures, trained without distillation. Hardware on the CC3220SF. $\Delta$: BeCNN minus ReLU, relative to ReLU (\%) for hardware and in pp for test acc.}
\label{tab:same-shape-board}
\centering\small\setlength{\tabcolsep}{2.5pt}
\renewcommand{\arraystretch}{1.3}
\begin{tabular*}{\textwidth}{@{\extracolsep{\fill}}llrrrrrr|rrr@{}}
\toprule & \multicolumn{1}{l}{} & \multicolumn{2}{c}{Flash (KiB)} & \multicolumn{2}{c}{RAM (KiB)} & \multicolumn{2}{c}{Latency (ms)} & \multicolumn{3}{c}{Test (\%)}\\\cmidrule(lr){3-4}\cmidrule(lr){5-6}\cmidrule(lr){7-8}\cmidrule(lr){9-11}
Dataset & Shape & BeCNN / ReLU & $\Delta$ & BeCNN / ReLU & $\Delta$ & BeCNN / ReLU & $\Delta$ & BeCNN & ReLU & $\Delta$\\\midrule
UCI-HAR & 1$\times$8, k3, s32 & 8.7 / 7.8 & +11.5\% & 2.4 / 2.4 & +1.2\% & 0.185 / 0.178 & +3.6\% & \textbf{80.00} $\pm$ 0.97 & 77.15 $\pm$ 0.51 & +2.85\\
UCI-HAR & 1$\times$8, k7, s16 & 9.1 / 8.2 & +11.0\% & 2.4 / 2.4 & +1.1\% & 0.294 / 0.286 & +2.8\% & \textbf{88.34} $\pm$ 0.58 & 87.12 $\pm$ 1.25 & +1.22\\
UCI-HAR & 1$\times$8, k9, s8 & 9.3 / 8.4 & +10.8\% & 2.5 / 2.4 & +1.1\% & 0.511 / 0.500 & +2.3\% & \textbf{89.91} $\pm$ 1.15 & 88.36 $\pm$ 0.86 & +1.55\\\midrule
HHAR & 1$\times$8, k3, s32, BN & 8.7 / 7.8 & +11.5\% & 2.4 / 2.4 & +1.2\% & 0.185 / 0.178 & +3.6\% & \textbf{73.71} $\pm$ 1.27 & 70.00 $\pm$ 0.14 & +3.70\\
HHAR & 3$\times$8, k5, s32, BN & 11.2 / 9.3 & +20.4\% & 2.5 / 2.5 & +1.1\% & 0.389 / 0.372 & +4.8\% & \textbf{85.19} $\pm$ 1.90 & 82.56 $\pm$ 0.79 & +2.63\\
HHAR & 3$\times$16, k7, s16, BN & 15.5 / 12.1 & +28.1\% & 2.6 / 2.6 & +1.0\% & 1.186 / 1.146 & +3.5\% & \textbf{91.63} $\pm$ 1.27 & 86.30 $\pm$ 0.94 & +5.32\\\midrule
PAMAP2 & 1$\times$8, k3, s32 & 10.4 / 9.5 & +9.3\% & 5.4 / 5.4 & +0.6\% & 0.448 / 0.446 & +0.5\% & \textbf{79.59} $\pm$ 0.28 & 74.75 $\pm$ 0.77 & +4.84\\
PAMAP2 & 1$\times$8, k3, s16, BN & 10.4 / 9.5 & +9.3\% & 5.4 / 5.4 & +0.6\% & 0.537 / 0.533 & +0.7\% & \textbf{82.42} $\pm$ 1.49 & 79.78 $\pm$ 1.91 & +2.64\\
PAMAP2 & 3$\times$8, k3, s8, BN & 13.5 / 11.6 & +16.2\% & 5.7 / 5.6 & +0.6\% & 1.532 / 1.511 & +1.4\% & 85.35 $\pm$ 1.15 & \textbf{85.38} $\pm$ 0.83 & $-$0.03\\
\bottomrule\end{tabular*}
\end{table*}

%% file: sections/16_results_projection.tex
\subsection{Projection and Recovery}
\label{sec:exp-projection}
\textbf{Projecting the distilled students.}
Table~\ref{tab:pipeline} projects each distilled student to powers of two, recovers it (Section~\ref{sec:projection}; recipe in Appendix~\ref{app:setup-details}) and runs it on the generated kernel (Section~\ref{sec:kernels}). ReLU uses conventional training in the style of TWN~\cite{twn}, which validation selects over our frozen-code recovery at eight of nine HAR models (Appendix~\ref{app:relu-baseline}). We therefore keep this stronger baseline for ReLU.

\textbf{Accuracy.} After projection, the BeCNN leads ReLU at 12 of 13 budgets, by up to 11.1\,pp. It loses 0.8\,pp of its dense accuracy on average, against 2.4\,pp for ReLU.

\textbf{Speed.} Projection makes both BeCNN and ReLU models about $2.2\times$ faster on the board, and up to $3.6\times$ on the DS-CNNs, so the BeCNN keeps its accuracy lead without losing the speedup.

\begin{table*}[!t]
\caption{Distilled students after powers-of-two projection, on the generated kernel, against each task's int8 teacher. $\Delta$: Proj.\ $-$ Dense; $\Delta_\text{teacher}$: Proj.\ $-$ Teacher (pp). Spd.: Dense/Proj. Bold: smaller drop.}
\label{tab:pipeline}
\centering\small\setlength{\tabcolsep}{2pt}\renewcommand{\arraystretch}{1.15}
\begin{tabular*}{\textwidth}{@{\extracolsep{\fill}}llrrrcrrrrrrcrrrr@{}}
\toprule
 & & Teacher & \multicolumn{7}{c}{\framework{} (Ours)} & \multicolumn{7}{c}{ReLU (TWN~\cite{twn})}\\
\cmidrule(lr){3-3}\cmidrule(lr){4-10}\cmidrule(l){11-17}
 & & Acc. & \multicolumn{4}{c}{Acc.\ (\%)} & \multicolumn{3}{c}{Lat.\ (ms)} & \multicolumn{4}{c}{Acc.\ (\%)} & \multicolumn{3}{c}{Lat.\ (ms)}\\
\cmidrule(lr){4-7}\cmidrule(lr){8-10}\cmidrule(lr){11-14}\cmidrule(l){15-17}
Dataset & Budget & (\%) & Dense & Proj. & $\Delta$ & $\Delta_\text{teacher}$ & Dense & Proj. & Spd. & Dense & Proj. & $\Delta$ & $\Delta_\text{teacher}$ & Dense & Proj. & Spd.\\
\midrule
\multirow{3}{*}{UCI-HAR}
 & 6k & \multirow{3}{*}{96.01} & 83.92 & 83.20 & $\mathbf{-0.72}$ & $\mathbf{-12.81}$ & 0.22 & 0.15 & 1.49$\times$ & 81.19 & 79.06 & $-2.13$ & $-16.95$ & 0.22 & 0.14 & 1.49$\times$\\
 & 15k &  & 92.29 & 92.10 & $\mathbf{-0.19}$ & $\mathbf{-3.91}$ & 0.51 & 0.25 & 2.08$\times$ & 89.97 & 88.61 & $-1.36$ & $-7.40$ & 0.42 & 0.24 & 1.73$\times$\\
 & 25k &  & 94.70 & 93.96 & $\mathbf{-0.74}$ & $\mathbf{-2.05}$ & 0.92 & 0.48 & 1.93$\times$ & 93.51 & 92.22 & $-1.29$ & $-3.79$ & 0.93 & 0.46 & 2.02$\times$\\
\midrule
\multirow{3}{*}{HHAR}
 & 7k & \multirow{3}{*}{94.72} & 81.55 & 81.15 & $\mathbf{-0.40}$ & $\mathbf{-13.57}$ & 0.26 & 0.15 & 1.66$\times$ & 77.02 & 70.08 & $-6.94$ & $-24.64$ & 0.25 & 0.15 & 1.66$\times$\\
 & 15k &  & 88.68 & 87.88 & $\mathbf{-0.80}$ & $\mathbf{-6.84}$ & 0.48 & 0.24 & 1.96$\times$ & 84.28 & 79.69 & $-4.59$ & $-15.03$ & 0.43 & 0.23 & 1.82$\times$\\
 & 60k &  & 92.91 & 92.16 & $\mathbf{-0.75}$ & $\mathbf{-2.56}$ & 2.49 & 1.12 & 2.22$\times$ & 92.70 & 88.43 & $-4.27$ & $-6.29$ & 1.98 & 0.87 & 2.27$\times$\\
\midrule
\multirow{3}{*}{PAMAP2}
 & 12k & \multirow{3}{*}{84.13} & 79.99 & 79.82 & $-0.17$ & $\mathbf{-4.31}$ & 0.45 & 0.34 & 1.32$\times$ & 77.42 & 77.84 & $\mathbf{+0.42}$ & $-6.29$ & 0.45 & 0.34 & 1.32$\times$\\
 & 20k &  & 85.92 & 85.35 & $\mathbf{-0.57}$ & $\mathbf{+1.22}$ & 0.71 & 0.44 & 1.62$\times$ & 85.18 & 82.70 & $-2.48$ & $-1.43$ & 0.75 & 0.47 & 1.61$\times$\\
 & 40k &  & 86.92 & 86.73 & $\mathbf{-0.19}$ & $\mathbf{+2.60}$ & 1.54 & 0.79 & 1.96$\times$ & 84.13 & 83.51 & $-0.62$ & $-0.62$ & 1.36 & 0.59 & 2.30$\times$\\
\midrule
\multirow{4}{*}{KWS}
 & 0.5M & \multirow{4}{*}{91.62} & 81.07 & 79.59 & $\mathbf{-1.48}$ & $\mathbf{-12.03}$ & 17.4 & 4.9 & 3.53$\times$ & 78.98 & 76.37 & $-2.61$ & $-15.25$ & 16.9 & 4.7 & 3.63$\times$\\
 & 1M &  & 86.41 & 84.51 & $\mathbf{-1.90}$ & $-7.11$ & 35.0 & 11.1 & 3.16$\times$ & 87.23 & 84.66 & $-2.57$ & $\mathbf{-6.96}$ & 36.5 & 11.9 & 3.08$\times$\\
 & 2M &  & 90.74 & 89.95 & $\mathbf{-0.79}$ & $\mathbf{-1.67}$ & 76.8 & 25.2 & 3.05$\times$ & 90.20 & 88.36 & $-1.84$ & $-3.26$ & 60.8 & 18.5 & 3.30$\times$\\
 & 4M &  & 92.60 & 90.62 & $-1.98$ & $\mathbf{-1.00}$ & 120.2 & 44.3 & 2.71$\times$ & 91.66 & 90.15 & $\mathbf{-1.51}$ & $-1.47$ & 117.1 & 35.9 & 3.26$\times$\\
\midrule
\multicolumn{2}{@{}l}{Mean} & & & & $\mathbf{-0.82}$ & $\mathbf{-4.93}$ & & & 2.21$\times$ & & & $-2.44$ & $-8.41$ & & & 2.27$\times$\\
\bottomrule
\end{tabular*}\end{table*}

\begin{table}[t]
\caption{UCI-HAR 25k BeCNN under each weight set. Flash and RAM in KiB, latency in ms. Speedup: against dense model.}
\label{tab:projection-board}
\centering\small\setlength{\tabcolsep}{2pt}
\begin{tabular*}{\columnwidth}{@{\extracolsep{\fill}}lrrrrrr@{}}
\toprule
Weights & Test (\%) & Instr. & Flash & RAM & Lat. & Speedup\\\midrule
Dense int8 & 94.70 & 22,819 & 11.9 & 2.5 & 0.919 & 1.00\\ \midrule
\multicolumn{7}{@{}l}{\emph{Ternary} $\{0,\pm1\}$}\\
\quad 70\% zeros & 92.06 & \textbf{9,121} & \textbf{6.9} & 2.4 & \textbf{0.352} & \textbf{2.61}\\
\quad 50\% zeros & 91.87 & 10,043 & 7.3 & 2.4 & 0.385 & 2.39\\
\quad 30\% zeros & 90.74 & 10,989 & 7.7 & 2.4 & 0.418 & 2.20\\
\quad Free & 93.20 & 9,710 & 7.2 & 2.4 & 0.378 & 2.43\\ \midrule
\multicolumn{7}{@{}l}{\emph{Powers of two} $\{0,\pm1,\pm2,\pm4\}$}\\
\quad 70\% zeros & 92.74 & 10,281 & 7.3 & 2.4 & 0.379 & 2.42\\
\quad 50\% zeros & 93.88 & 11,990 & 8.0 & 2.5 & 0.432 & 2.13\\
\quad 30\% zeros & 93.42 & 13,538 & 8.8 & 2.5 & 0.484 & 1.90\\
\quad Free & \textbf{93.96} & 13,433 & 8.6 & 2.5 & 0.477 & 1.93\\
\bottomrule\end{tabular*}
\end{table}

\textbf{Choice of weight set.}
\label{sec:projection-sets}
Table~\ref{tab:projection-board} compares the weight sets on one student. Ternary at 70\% zeros is the fastest ($2.61\times$) but loses 2.6\,pp. Powers of two with a free share lose 0.7\,pp and still run $1.93\times$ faster.

\textbf{Which parameters recover the projection loss.}
\label{sec:recovery-ablation}
Table~\ref{tab:recovery-controls} isolates the Bernstein coefficients. At ternary codes, where projection loses the most, training only the coefficients gives 75 to 96\% of the full recovery in seven of eight cases, and up to 17.8\,pp at 85\% zeros. The activations therefore carry the recovery; biases, scales and the head add the rest.

\begin{table}[t]\caption{Recovery ablation on 16-filter BeCNNs (dense: 92.00\% on UCI-HAR, 89.53\% on HHAR). None: no recovery; Coef.: Bernstein coefficients only; Full: coefficients, biases, scales and head. Test (\%).}
\label{tab:recovery-controls}
\centering\small\setlength{\tabcolsep}{2pt}
\begin{tabular*}{\columnwidth}{@{\extracolsep{\fill}}lrrrrrr@{}}\toprule & \multicolumn{3}{c}{UCI-HAR} & \multicolumn{3}{c}{HHAR}\\\cmidrule(lr){2-4}\cmidrule(lr){5-7}
Ternary codes & None & Coef. & Full & None & Coef. & Full\\\midrule
30\% zeros & 87.69 & 88.99 & 89.31 & 76.84 & 78.20 & 80.79\\
50\% zeros & 87.34 & 90.45 & 90.78 & 74.04 & 78.25 & 79.57\\
70\% zeros & 85.21 & 90.07 & 90.76 & 79.82 & 82.83 & 82.96\\
85\% zeros & 68.57 & 86.40 & 88.51 & 68.37 & 78.62 & 81.95\\
\bottomrule\end{tabular*}\end{table}

%% file: sections/17_results_kernels.tex
\subsection{Kernel Evaluation}
\label{sec:exp-kernels}

\textbf{TFLM+CMSIS-NN vs.\ \framework{}'s kernel.}
\label{sec:exp-kernel-runtime}
Table~\ref{tab:kernel-runtime} checks that our ReLU and GELU baselines are strong by running the same trained network on TFLM with CMSIS-NN and on our dense kernel. TFLM's int8 models come from the TFLite Converter, as for the teachers, so they share our float32 weights but not our int8 bits. Our kernel beats TFLM+CMSIS-NN on every metric: 0.6 to 1.8\,pp more accurate, $1.26$--$2.66\times$ faster, with up to $10.8\times$ less flash and $3.4\times$ less RAM. Our ReLU and GELU baselines are therefore stronger than the standard deployment, and \framework{}'s lead over them does not come from a weak kernel. The Bernstein network of the same shape keeps that speedup ($1.22$--$2.57\times$) and is up to 5.3\,pp more accurate than the ReLU network on our kernel.

\begin{table*}[!t]
\caption{The same network on TFLM+CMSIS-NN~\cite{tflm,cmsisnn} and on our dense kernel (Ours), with \framework{}$_\text{dense}$ of the same shape on ours. TFLM's KWT uses one custom GELU operator. Speedup: against TFLM+CMSIS-NN.}
\label{tab:kernel-runtime}
\centering\small\setlength{\tabcolsep}{3pt}
\begin{tabular*}{\textwidth}{@{\extracolsep{\fill}}lllrrrrrr@{}}
\toprule Network & Model & Kernel & Test (\%) & Instr. & Lat.\ (ms) & Flash (KiB) & RAM (KiB) & Speedup\\\midrule
HHAR CNN, 3$\times$16, k7, s16 & ReLU & TFLM+CMSIS-NN & $84.50\pm0.51$ & 79,406 & 3.05 & 70.9 & 6.8 & --\\
 & ReLU & Ours & $86.30\pm0.94$ & 28,250 & 1.15 & 12.1 & 2.6 & 2.66$\times$\\
 & \framework{}$_\text{dense}$ & Ours & $\mathbf{91.63}\pm1.27$ & 28,184 & 1.19 & 15.5 & 2.6 & 2.57$\times$\\
\midrule
DS-CNN, 0.5M & ReLU & TFLM+CMSIS-NN & $78.36\pm0.77$ & 674,407 & 21.33 & 84.0 & 9.1 & --\\
 & ReLU & Ours & $78.98\pm0.54$ & 438,713 & 16.90 & 7.8 & 2.7 & 1.26$\times$\\
 & \framework{}$_\text{dense}$ & Ours & $\mathbf{81.07}\pm0.35$ & 423,591 & 17.42 & 8.0 & 2.7 & 1.22$\times$\\
\midrule
KWT, 3$\times$256 & GELU & TFLM+CMSIS-NN & $91.82\pm0.25$ & 79.1\,M & 2,698.1 & 282.0 & 89.4 & --\\
 & GELU & Ours & $92.43\pm0.37$ & 40.5\,M & 1,464.5 & 213.7 & 60.5 & 1.84$\times$\\
 & \framework{}$_\text{dense}$ & Ours & $\mathbf{94.83}\pm0.41$ & 38.6\,M & 1,414.7 & 257.1 & 60.5 & 1.91$\times$\\
\bottomrule\end{tabular*}
\end{table*}

\begin{figure}[t]
\centering\includegraphics[width=\columnwidth]{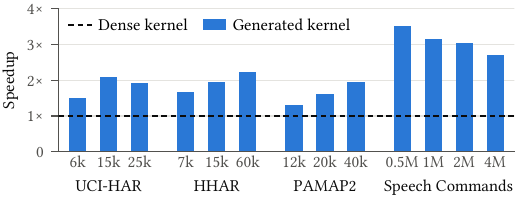}
\caption{Board speedup of the projected BeCNNs over their unprojected models. Projected weights only pay off on the generated kernel: on the dense kernel, the latency stays the same ($1\times$).}
\label{fig:kernel-generated}
\end{figure}

\textbf{Projected codes: dense vs.\ generated kernel.}
\label{sec:exp-kernel-generated}
On the dense kernel, projected codes take exactly the unprojected model's cycles on all 13 BeCNNs, because it still multiplies every weight. Only the generated kernel turns them into speed, $1.3$--$3.5\times$ (Fig.~\ref{fig:kernel-generated}), so projection and the generated kernel pay off only together.

\textbf{Weights as code vs.\ as data.}
\label{sec:exp-storage}
Table~\ref{tab:kernel-storage} stores the same codes three ways (Section~\ref{sec:gen-kernel}). On the transformer, code is $1.79\times$ faster than dense but needs $1.6\times$ its flash, while data cuts the flash by a third and is barely faster ($1.03\times$). The mix is $1.22\times$ faster within the unprojected model's flash. On the CNN, only code pays off ($2.22\times$ for 9\% more flash). Data runs $2.5\times$ slower: each nonzero code costs three instructions (an index load, an input load and an add), while one multiply-accumulate instruction of the dense kernel handles two weights.

\begin{table}[t]
\caption{A projected KWT (6$\times$64) and BeCNN (HHAR 60k) with their codes stored three ways: Code, Data and Mix. Dense: the unprojected model on the dense kernel. Speedup: over Dense.}
\label{tab:kernel-storage}
\centering\small\setlength{\tabcolsep}{2pt}\renewcommand{\arraystretch}{1.2}
\begin{tabular*}{\columnwidth}{@{\extracolsep{\fill}}lrrrrrr@{}}
\toprule & \multicolumn{3}{c}{KWT} & \multicolumn{3}{c}{HHAR}\\\cmidrule(lr){2-4}\cmidrule(lr){5-7}
Weights & Lat.\ (ms) & Speedup & Flash (KiB) & Lat.\ (ms) & Speedup & Flash (KiB)\\\midrule
Dense &  1,859 & 1.00$\times$ & 249.5 & 2.49 & 1.00$\times$ & 16.6\\
Code & 1,036 & \textbf{1.79}$\times$ & 391.3 & 1.12 & \textbf{2.22}$\times$ & 18.1\\
Data & 1,810 & 1.03$\times$ & \textbf{167.9} & 6.14 & 0.41$\times$ & \textbf{10.2}\\
Mix & 1,529 & 1.22$\times$ & 246.3 & 2.79 & 0.89$\times$ & 15.9\\
\bottomrule\end{tabular*}
\end{table}

%% file: sections/18_results_transformer_mlp.tex
\subsection{Beyond CNNs: Transformers and MLPs}
\label{sec:exp-extensions}
The same pipeline carries over to two other model families: keyword-spotting transformers and Wi-Fi localization MLPs.

\subsubsection{\textbf{Keyword transformers}}
\label{sec:exp-kwt}
We replace the GELU feed-forward (FFN) activation of a KWT-1-style transformer~\cite{berg2021kwt} with one degree-15 Bernstein curve per hidden unit. Students have embedding width 64, one attention head, six or three blocks and FFN width $h\in\{256,128,64\}$, and are distilled ($T=3$) from the AST teacher (Section~\ref{sec:setup}) through a separate distillation token and head, as in DeiT~\cite{touvron2021deit}, with KWT's training recipe. GELU becomes one 256-entry table shared by all units of a layer, and each Bernstein unit its own 64-entry table (Section~\ref{sec:lut}); both run on \framework{}'s kernels.

\textbf{Results.} Table~\ref{tab:kwt} compares GELU with Bernstein curves, before and after projection to powers of two with 70\% zeros. \framework{}$_\text{dense}$ beats GELU at every shape, and the gain grows with compression, up to 2.40\pp{} at three blocks. It needs 1.9--4.8\% fewer instructions and runs 1.2--3.5\% faster at the same RAM, for 9--22\% more flash. Bernstein activations also let the FFN halve its width: \framework{}$_\text{dense}$ at $h=128$ beats GELU at $h=256$ with 25\% lower latency and 16\% less flash.

Projection costs \framework{} at most 0.9\pp{} and speeds it up $1.2$--$1.4\times$. The attention products limit the speedup: they multiply two activations, have no weights to project, and make up 20--34\% of the unprojected model's multiply-accumulates.

\begin{table}[t]
\caption{KWTs on Speech Commands (35 words). Shape: blocks\,$\times$\,FFN width $h$. $^\ast$Does not fit the board: measured in QEMU on TFLM's reference kernels, so it has no latency.}
\label{tab:kwt}
\centering\footnotesize\setlength{\tabcolsep}{1.5pt}\renewcommand{\arraystretch}{1.3}
\begin{tabular*}{\columnwidth}{@{\extracolsep{\fill}}llrrrrr@{}}
\toprule
Shape & Model & Test (\%) & Instr.\ (M) & Lat.\ (ms) & Flash (KiB) & RAM (KiB)\\\midrule
12$\times$3072 & AST$^\ast$ & 97.10 & 95,738 & --- & 83,797.0 & 2,353.9\\\midrule
\multirow{3}{*}{6$\times$256} & GELU & $96.06\pm0.13$ & 80.4 & 2,908 & 386.0 & 60.5\\
 & \framework{}$_\text{dense}$ & $\mathbf{96.48}\pm0.31$ & 76.5 & 2,807 & 471.2 & 60.5\\
 & \framework{}$_\text{projected}$ & $95.90\pm0.21$ & 51.3 & 1,995 & 468.0 & 62.5\\\midrule
\multirow{3}{*}{6$\times$128} & GELU & $95.33\pm0.18$ & 62.1 & 2,224 & 281.0 & 60.5\\
 & \framework{}$_\text{dense}$ & $\mathbf{96.21}\pm0.13$ & 60.2 & 2,176 & 323.8 & 60.5\\
 & \framework{}$_\text{projected}$ & $95.34\pm0.23$ & 43.6 & 1,655 & 320.0 & 62.5\\\midrule
\multirow{3}{*}{6$\times$64} & GELU & $94.77\pm0.13$ & 52.9 & 1,882 & 228.7 & 60.5\\
 & \framework{}$_\text{dense}$ & $\mathbf{95.68}\pm0.11$ & 51.9 & 1,859 & 249.5 & 60.5\\
 & \framework{}$_\text{projected}$ & $95.44\pm0.21$ & 40.6 & 1,529 & 246.3 & 62.5\\\midrule
\multirow{3}{*}{3$\times$256} & GELU & $92.43\pm0.37$ & 40.5 & 1,464 & 213.7 & 60.5\\
 & \framework{}$_\text{dense}$ & $\mathbf{94.83}\pm0.41$ & 38.6 & 1,415 & 257.1 & 60.5\\
 & \framework{}$_\text{projected}$ & $94.51\pm0.08$ & 26.5 & 1,053 & 242.2 & 62.5\\
\bottomrule\end{tabular*}
\end{table}

\subsubsection{\textbf{MLPs for Wi-Fi localization}}
\label{sec:mlp-results}
UJIIndoorLoc~\cite{torressospedra2014uji} maps Wi-Fi RSSI values to a building and a floor; we report joint building-and-floor accuracy, the share of samples whose building and floor are both correct. The teacher is Kim et al.'s SAE-pretrained ReLU MLP~\cite{kim2018indoor}, 520--256--128--64--128, with two softmax heads instead of their multi-label output; it reaches $92.02\pm0.57\%$ float32 joint accuracy, against the published 91.18\%. The students have a single Bernstein hidden layer of 128, 64 or 32 units, are distilled and then projected to ternary weights with 70\% zeros.

\textbf{Results.} One Bernstein hidden layer replaces the teacher's four (Table~\ref{tab:mlp-onelayer}): the 64-unit student reaches 92.59\% joint accuracy against 92.02\%. Ternary projection with 70\% zeros costs at most 0.09\pp{} and makes the students 12 to $48\times$ faster than the teacher on TFLM, with up to $4.9\times$ less flash and $2.7\times$ less RAM.

\begin{table}[t]
\centering
\caption{MLPs on UJIIndoorLoc: the teacher on TFLM with CMSIS-NN against \framework{} students; ($\downarrow\times$): teacher / student. Test: joint building-and-floor accuracy.}
\label{tab:mlp-onelayer}
\centering\small\setlength{\tabcolsep}{1pt}\renewcommand{\arraystretch}{1.36}
\begin{tabular*}{\columnwidth}{@{\extracolsep{\fill}}llrrrr@{}}
\toprule
Shape & Model & Test (\%) & Lat.\ (ms) & Flash (KiB) & RAM (KiB)\\\midrule
\begin{tabular}[t]{@{}l@{}}520--256--128\\--64--128--8\end{tabular} & Teacher & $92.02\pm0.64$ & 14.68 & 211.4 & 2.9\\\midrule
\multirow{2}{*}{520--128--8} & \framework{}$_\text{dense}$ & $92.41\pm0.40$ & \rd{6.66}{2.21} & \rd{162.2}{1.30} & \rd{2.4}{1.19}\\
 & \framework{}$_\text{projected}$ & $92.41\pm0.52$ & \rd{1.18}{12.5} & \rd{108.8}{1.94} & \rd{1.5}{1.98}\\\midrule
\multirow{2}{*}{520--64--8} & \framework{}$_\text{dense}$ & $92.59\pm0.36$ & \rd{3.34}{4.40} & \rd{91.9}{2.30} & \rd{2.4}{1.19}\\
 & \framework{}$_\text{projected}$ & $92.56\pm0.04$ & \rd{0.60}{24.4} & \rd{65.4}{3.23} & \rd{1.2}{2.41}\\\midrule
\multirow{2}{*}{520--32--8} & \framework{}$_\text{dense}$ & $91.39\pm0.40$ & \rd{1.68}{8.74} & \rd{56.7}{3.73} & \rd{2.4}{1.19}\\
 & \framework{}$_\text{projected}$ & $91.30\pm1.42$ & \rd{0.31}{47.6} & \rd{43.4}{4.87} & \rd{1.1}{2.74}\\
\bottomrule
\end{tabular*}
\end{table}

%% file: sections/20_discussion.tex
\section{Limitations and Future Work}
\paragraph{\textbf{Remaining multiplications.}} The int8 classifier still multiplies.
In the keyword transformers, the attention products $QK^{\top}$ and softmax$\cdot V$
multiply two activations, so there are no weights to project. These products are
20--34\% of the unprojected model's multiply-accumulates, which limits the speedup.
Multiplication-free or lookup-based attention is a natural next step.

\paragraph{\textbf{On-device adaptation.}} The projected codes are frozen, and training only the Bernstein coefficients
recovers most of the projection loss (Section~\ref{sec:recovery-ablation}).
This suggests updating only the curves on the device, for example to adapt to a
new user or to sensor drift, while the weight code stays fixed in flash.

\paragraph{\textbf{Energy and hardware coverage.}}
We report latency, flash and RAM on a single Cortex-M4 board.
The RISC-V results without a hardware multiplier come from emulation.
Future work will measure energy per inference on the board. It will also run \framework{} on
multiplier-less RISC-V silicon and on Cortex-M cores with vector extensions,
where the trade-off between projected and dense weights may change.

%% file: sections/21_conclusion.tex
\section{Conclusion}
We presented \framework{}, an end-to-end framework that shifts model capacity from the weights to learnable Bernstein activations, so that compact MCU models keep their accuracy while their convolutions become multiplication-free.
Across six sensing tasks, BeCNNs outperform ReLU CNNs of identical architecture, lose little accuracy after weight projection, and run faster than strong MCU baselines, with gains that grow on cores without a hardware multiplier.
These results suggest that learnable activations are an effective place to put model capacity when weights must be small and cheap.

%% file: sections/22_appendix.tex
\section{Dataset and Build Details}
\label{app:setup-details}
\paragraph{\textbf{Dataset preprocessing.}} HAR windows are 128 samples at 50\,Hz (2.56\,s, 50\% overlap), lie within single-label runs, and are standardized per channel on training statistics. UCI-HAR keeps total acceleration and body gyroscope. HHAR uses the accelerometer and gyroscope of the six phones that record both, resampled to 50\,Hz, without the null label. PAMAP2 uses the $\pm16$\,g accelerometer and gyroscope of its three IMUs, with missing samples interpolated, resampled from 100\,Hz, without the transient label~0. The DS-CNNs' 12 Speech Commands classes are ten words, silence and unknown, with 30\,ms frames, 20\,ms hop, 40 mel bins and 10 coefficients~\cite{mlperftiny}. UJIIndoorLoc splits its official training file 70:30~\cite{kim2018indoor} and tests on its official validation file. 

\paragraph{\textbf{Teacher models.}} Table~\ref{tab:teacher-details} lists every teacher. ResNet, DeepSense and DeepConvLSTM are trained on our splits with their published recipes: Adam at $10^{-3}$, halved on plateau, batch 64 (ResNet~\cite{wang2017resnet,ismailfawaz2019review}); Adam at $10^{-4}$, $\beta=(0.5,0.9)$, L2 $5\times10^{-4}$, batch 64 (DeepSense~\cite{yao2017deepsense}); RMSProp at $10^{-3}$, dropout 0.5, batch 100 (DeepConvLSTM~\cite{ordonez2016deepconvlstm}). On top of these, they use the students' augmentation and train for at most 500 epochs, with early stopping on validation accuracy (patience 100). DeepSense splits each 128-sample window into 8 intervals of 16 samples, because our windows are shorter than the published 5\,s. The DS-CNN is the released MLPerf Tiny model~\cite{mlperftiny,mlperfkws}, and its int8 network is the released TFLite file. AST is the released checkpoint \texttt{MIT/ast-finetuned-speech-commands-v2}~\cite{gong2021ast}, which we convert to int8 with the TFLite Converter on 500 training clips. The UJIIndoorLoc teacher follows Kim et al.~\cite{kim2018indoor}: a stacked autoencoder (256--128--256) whose encoder feeds classifier layers 64--128, trained with Adam, dropout 0.2, batch 10 for 20 epochs.

\begin{table}[H]
\caption{Teachers. Float32 and int8: test accuracy (\%); UJIIndoorLoc: joint building-and-floor accuracy. $\pm$: population SD over three seeds; no $\pm$: one released model.}
\label{tab:teacher-details}
\centering\footnotesize\setlength{\tabcolsep}{2pt}
\begin{tabular*}{\columnwidth}{@{\extracolsep{\fill}}llrrr@{}}\toprule
Task & Teacher & Params & Float32 & Int8\\\midrule
UCI-HAR & ResNet~\cite{wang2017resnet} & 0.51\,M & 96.52\,$\pm$\,0.46 & 96.01\,$\pm$\,0.32\\
HHAR & DeepSense~\cite{yao2017deepsense} & 0.47\,M & 95.08\,$\pm$\,0.52 & 94.72\,$\pm$\,0.57\\
PAMAP2 & DeepConvLSTM~\cite{ordonez2016deepconvlstm} & 0.85\,M & 86.99\,$\pm$\,1.10 & 84.13\,$\pm$\,1.86\\
KWS (12) & DS-CNN~\cite{mlperftiny,mlperfkws} & 24.9\,K & 92.17 & 91.62\\
KWS (35) & AST~\cite{gong2021ast} & 85.4\,M & 98.03 & 97.10\\
UJIIndoorLoc & MLP~\cite{kim2018indoor} & 0.18\,M & 92.02\,$\pm$\,0.57 & 92.02\,$\pm$\,0.64\\
\bottomrule\end{tabular*}
\end{table}

\paragraph{\textbf{Build and measurement.}} Images are compiled with GNU Arm GCC 15.2.1 (\texttt{-O2}, \texttt{-mcpu=cortex-m4}, soft-float ABI); TFLite Micro with CMSIS-NN~\cite{tflm,cmsisnn} uses its release build. Images are loaded over JTAG, and test inputs are streamed over UART.

\paragraph{\textbf{Post-projection recovery training.}} Recovery after projection uses Adam with a cosine schedule, batch size 256, and $\alpha=0.5$, $T=2$ in Eq.~\eqref{eq:kd}, for 50 epochs on HAR and 100 on the DS-CNNs. BeCNNs with BatchNorm keep it live during recovery, add the out-of-bounds penalty of Eq.~\eqref{eq:ob}, and fold it into the codes' scales and biases at the end.

\section{TFLM Reference Kernels}
\label{app:tflm_base}
Table~\ref{tab:tflm-ref} repeats the int8 teachers on TFLM's default reference kernels, in plain C, with no optimized kernels. They run 2.5 to 6.1$\times$ slower than with CMSIS-NN, with 5 to 31\% less flash and about the same RAM. AST has no CMSIS-NN build and does not fit the board, so Table~\ref{tab:e2e} reports it on these reference kernels in QEMU, and it is not listed here.

\begin{table}[H]
\caption{Int8 teachers on TFLM's reference kernels against TFLM with CMSIS-NN, seed 0.}
\label{tab:tflm-ref}
\centering\footnotesize\setlength{\tabcolsep}{2pt}
\begin{tabular*}{\columnwidth}{@{\extracolsep{\fill}}lrrrr@{}}\toprule
Teacher & Lat. (ms) & Slower & Flash (KiB) & RAM (KiB)\\\midrule
ResNet & 22,205 & 4.9$\times$ & 576.1 & 77.1\\
DeepSense & 2,373 & 2.5$\times$ & 625.3 & 51.0\\
MLPerf Tiny DS-CNN & 1,591 & 6.1$\times$ & 92.9 & 23.5\\
UJIIndoorLoc MLP & 60.8 & 4.1$\times$ & 200.3 & 2.38\\
\bottomrule\end{tabular*}
\end{table}

\begin{figure*}[t]\centering\includegraphics[width=\textwidth]{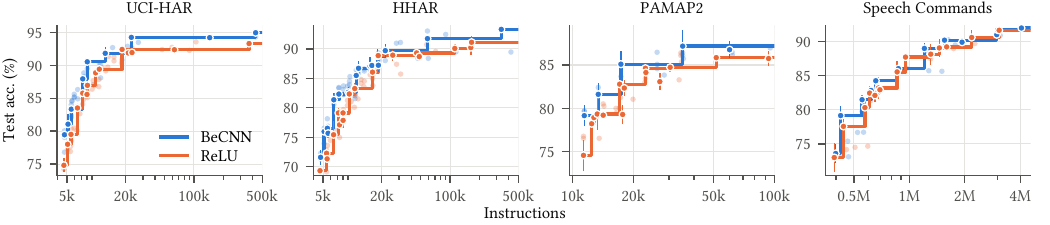}
\Description{Four plots, one each for UCI-HAR, HHAR, PAMAP2 and Speech Commands, show int8 test accuracy against dense int8 kernel instructions on a log axis for the architecture-search fronts of BeCNNs and ReLU models. On the three HAR datasets the BeCNN staircase lies above the ReLU staircase at most instruction counts, with the largest gaps at the lowest counts. On Speech Commands it lies above the ReLU staircase below about one million instructions; above that the two cross and run close together up to 92 percent.}
\caption{Pareto fronts: candidates are selected on three-seed validation accuracy and plotted at their int8 test accuracy. Markers: front models; step line: best test accuracy up to each instruction count; faint dots: all three-seed candidates.}
\label{fig:nas-fronts}\end{figure*}

\section{Multiply-Extension Control}
\label{app:rv32emc}
Table~\ref{tab:rv32emc} repeats the instruction counts of Table~\ref{tab:e2e-nomul} on the same RISC-V core with the multiply extension (RV32EMC). With a hardware multiplier, the reductions return to the same order of magnitude as on the Cortex-M4, so the larger reductions on RV32EC come from the missing multiplier, not from RISC-V.

\begin{table}[H]
\caption{Teacher and projected student on RV32EMC: QEMU instructions per inference, seed 0. Reduction: teacher / student, with the Cortex-M4 and RV32EC reductions of Table~\ref{tab:e2e-nomul} for comparison.}
\label{tab:rv32emc}
\centering\footnotesize\setlength{\tabcolsep}{2pt}
\begin{tabular*}{\columnwidth}{@{\extracolsep{\fill}}lrrrrr@{}}\toprule
 & \multicolumn{2}{c}{RV32EMC (M instr.)} & \multicolumn{3}{c}{Reduction ($\times$)}\\
\cmidrule(lr){2-3}\cmidrule(l){4-6}
Task & Teacher & Student & M4 & RV32EMC & RV32EC\\\midrule
UCI-HAR & 374.5 & 0.065 & 8,962 & 5,760 & 54,798\\
HHAR & 68.9 & 0.237 & 767 & 291 & 2,890\\
PAMAP2 & 1,225.1 & 0.138 & --- & 8,879 & 154,680\\
KWS & 18.9 & 2.398 & 5.2 & 7.9 & 147\\
\bottomrule\end{tabular*}
\end{table}

\section{Neural Architecture Search Details}
\label{app:nas}
The HAR and Speech Commands searches took 11.1 and 33.0 GPU-hours, including pruned trials. Candidates of non-dominated rank $\le2$ are evaluated over three seeds: 163 BeCNN and 166 ReLU models.

Figure~\ref{fig:nas-fronts} shows the resulting fronts. On all three HAR datasets, the BeCNN envelope lies above the ReLU envelope at most instruction counts, with the largest gaps at the smallest models. It peaks at 94.93\% against 93.29\% on UCI-HAR, 93.32\% against 91.44\% on HHAR and 87.18\% against 85.85\% on PAMAP2. On HHAR and PAMAP2 the BeCNN peak needs fewer instructions. On Speech Commands, the BeCNN leads almost everywhere below 0.9M instructions, by up to 6.13\,pp. Between 0.9M and 3M the two alternate, and above 3M the BeCNN leads by at most 1.09\,pp.

\section{ReLU Projection Baseline}
\label{app:relu-baseline}
Table~\ref{tab:relu-projection} compares two ways to recover the ReLU students after projection: conventional TWN-style training, which retrains the weights, and our frozen-code pass, which trains only the scales, biases and head. We choose the method on validation data. The table reports int8 validation accuracy, and its numbers differ from the test accuracies of Table~\ref{tab:pipeline}. Conventional training wins at eight of nine budgets, so the ReLU baseline uses it.

\begin{table}[H]
\caption{ReLU students of Table~\ref{tab:pipeline} at powers of two $\{0,\pm1,\pm2,\pm4\}$ with a free share of zeros: TWN-style training~\cite{twn} (Conventional) against our frozen-code single pass (BeFit). int8 validation accuracy (\%), three-seed mean. Bold: better of the pair.}
\label{tab:relu-projection}
\centering\footnotesize\setlength{\tabcolsep}{2pt}
\begin{tabular*}{\columnwidth}{@{\extracolsep{\fill}}llrr@{}}
\toprule
Dataset & Budget & TWN~\cite{twn}  & BeFit\\\midrule
UCI-HAR & 6k  & \textbf{87.23} & 87.14\\
        & 15k & 93.45 & \textbf{93.54}\\
        & 25k & \textbf{95.13} & 93.78\\\midrule
HHAR    & 7k  & \textbf{80.15} & 78.02\\
        & 15k & \textbf{85.27} & 84.86\\
        & 60k & \textbf{94.21} & 91.59\\\midrule
PAMAP2  & 12k        & \textbf{77.90} & 76.62\\
        & 20k        & \textbf{83.86} & 82.61\\
        & 40k & \textbf{83.23} & 80.24\\
\bottomrule\end{tabular*}
\end{table}